\documentclass[10pt,twocolumn,letterpaper]{article}

\usepackage{titling}
\usepackage{iccv}
\usepackage{times}
\usepackage{epsfig}
\usepackage{graphicx}
\usepackage{amsmath}
\usepackage{amssymb}
\usepackage{multirow}

\usepackage{graphics}
\usepackage{pifont}
\usepackage{color}
\usepackage{booktabs}
\usepackage{multirow}
\usepackage{subcaption}
\usepackage{gensymb}
\usepackage{bm}
\usepackage[table]{xcolor}
\usepackage{soul}
\usepackage{amssymb}

\usepackage[breaklinks=true,bookmarks=false]{hyperref}
\iccvfinalcopy 

\def\iccvPaperID{2740} 

\ificcvfinal\fi
\begin{document}
\newcommand{\raquel}[1]{\textcolor{black}{#1}}
\newcommand{\shenlong}[1]{\textcolor{black}{#1}}
\newcommand{\weichiu}[1]{\textcolor{black}{#1}}
\newcommand{\shivam}[1]{\textcolor{black}{#1}}
\newcommand{\todo}[1]{\textcolor{black}{#1}}
\newcommand{\slwang}[1]{\textcolor{black}{#1}}
\title{DeepPruner: Learning Efficient Stereo Matching via Differentiable PatchMatch}
\author{Shivam Duggal \thanks{Work done as part of the Uber AI Residency program.} $^{1}$, Shenlong Wang$^{1,2}$, Wei-Chiu Ma$^{1,3}$, Rui Hu$^{1}$ and Raquel Urtasun$^{1,2}$\\
{$^1$Uber ATG, $^2$University of Toronto, $^3$Massachusetts Institute of Technology}
}

\maketitle
\thispagestyle{empty}

\begin{abstract}
Our goal is to significantly speed up the runtime of current state-of-the-art  stereo algorithms to enable real-time inference. 
Towards this goal, we developed a differentiable PatchMatch module that allows us to discard most disparities without requiring full cost volume evaluation. 
We then exploit this representation to learn which range to prune for each pixel.
 By progressively reducing the search space and effectively propagating such information, we are able to efficiently compute the cost volume for high likelihood hypotheses and achieve savings in both memory and computation.
Finally, an image guided refinement module is exploited to further improve the performance. Since all our components are differentiable, the full network can be trained end-to-end. 
Our experiments show that our method achieves competitive results on KITTI and SceneFlow datasets while running in real-time at 62ms.

\end{abstract}

\section{Introduction}

Stereo estimation is the process of estimating depth (or disparity) from a pair of images with overlapping fields of view. 
It is a fundamental building block for many applications such as robotics and computational photography. 
Despite many decades of research, stereo estimation of real-world scenes remains an open problem. 
\weichiu{State-of-the-art approaches still have difficulties tackling repetitive structures, texture-less regions, occlusions and thin objects. 
Furthermore, runtime is also a challenge.
While real-time inference is required for many applications, it is hard to achieve \slwang{without a significant compromise on accuracy.}}

Classic stereo approaches typically start by computing robust feature representations \cite{ncc, sad,  daisy, sift, zbontar2016stereo}. A cost volume is then computed for  each pixel,  encoding the similarity between its representation and the representation of all pixels along the corresponding epipolar line on the other image. Post-processing techniques \cite{hirschmuller2008stereo,szeliski2008comparative} are typically exploited
for smoothing and noise removal. The lowest-cost disparity value is then picked for each pixel as the final prediction. 

Yet this pipeline is usually computationally demanding, due to the size of the solution space  and the use of sophisticated post-processing procedures. 
	A plethora of acceleration approaches have been proposed toward achieving real-time performance, such as solution space pruning \cite{bleyer2011patchmatch}, coarse-to-fine cost volume computation \cite{khamis2018stereonet} and efficient variational optimization \cite{ferstl2013image}. \weichiu{Among them, PatchMatch \cite{barnes2009patchmatch} is one of the most popular techniques used to achieve competitive performance with real-time inference \cite{besse2014pmbp,bleyer2011patchmatch}.}

\begin{figure*}
\includegraphics[width=0.99\linewidth, trim={5mm 70mm 26mm 0mm},clip]{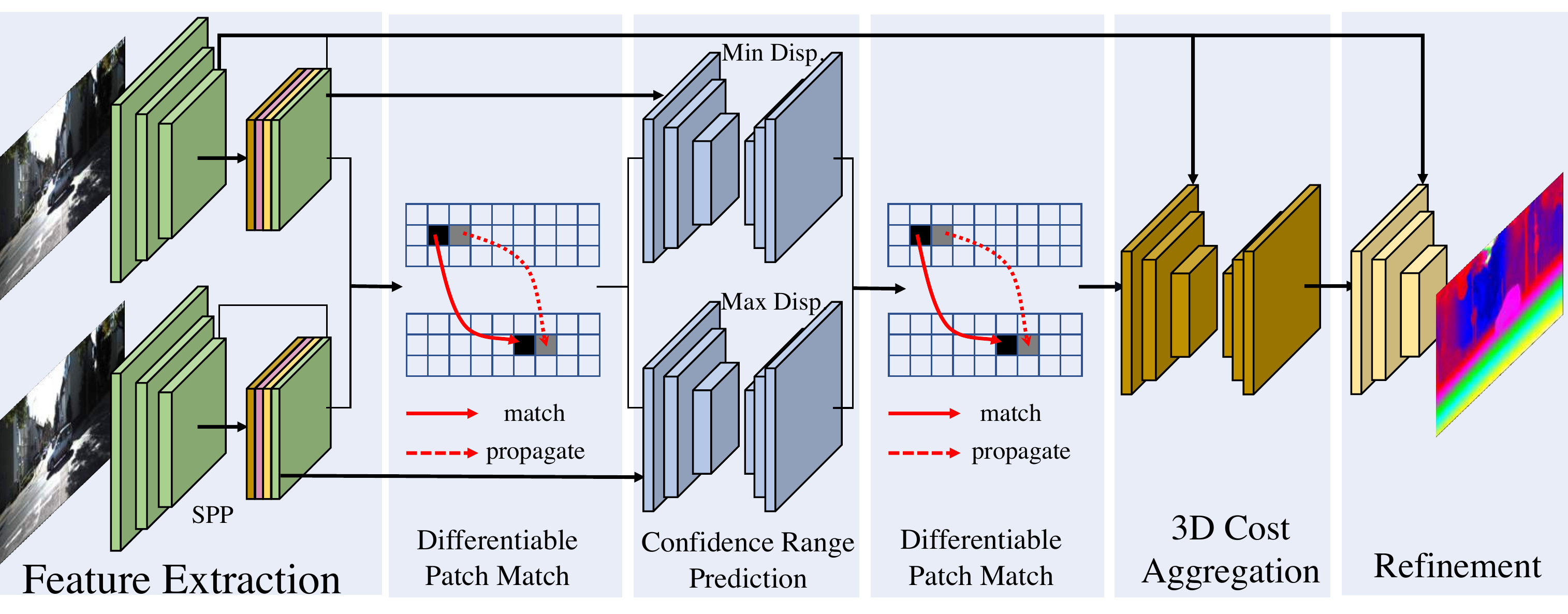}
\caption{\textbf{Overview:}  \weichiu{Given a pair of stereo images, we first extract deep multi-scale features. Then we exploit differentiable PatchMatch to estimate a small subset of disparities for each pixel and capitalize on confidence range predictor to further prune out the solution space. Unlike other approaches \cite{chang2018pyramid,kendall2017end} which operate on the entire disparity search range, we only aggregate the cost within the reduced search range. Finally, we leverage a light-weight network to refine the stereo output.}}

\label{fig:overview}
\end{figure*}

More recently, with the flourishing of deep learning, networks inspired by traditional stereo matching pipelines have been designed. They achieve state-of-the-art results by learning deep representations \cite{luo2016efficient,mayer2016large} and building deep cost volumes \cite{kendall2017end,chang2018pyramid, khamis2018stereonet}.  
\weichiu{While these approaches can be accelerated with GPUs, memory consumption and computation requirement is still a concern for most architectures.}
An alternative is to design a regression network that takes a stereo pair as input and directly regresses disparity without explicit matching or cost volume construction. Unfortunately, there is typically a large performance gap between real-time methods and best-performing algorithms. 

Our goal is to significantly speed up the runtime of current state-of-the-art deep stereo algorithms to enable real-time inference. We build our model based on two key observations: first, the search space for stereo matching is large, yet a lot of candidates can be confidently discarded without requiring full evaluation; second, due to the coherent nature of the world, adjacent pixels often possess similar disparities. \weichiu{This suggests that once we know the disparity of one pixel, we can effectively propagate such information to its neighbors.}

With these intuitions in mind, we propose DeepPruner, a real-time stereo matching model. Specifically, we first leverage a novel  differentiable PatchMatch algorithm to obtain a sparse representation of the cost volume. 
We then exploit this representation to learn which range to prune for each pixel.
Finally, an image guided refinement module is exploited to further improve the performance. Since all components are differentiable, the full network can be trained in an end-to-end fashion.  \weichiu{By progressively reducing the search space and effectively propagating such information, we are able to efficiently compute the cost volume for high likelihood hypotheses and significantly reduces both memory consumption and computational cost.}

\weichiu{We demonstrate the \slwang{efficiency and} effectiveness of our approach on the challenging SceneFlow \cite{mayer2016large} and KITTI \cite{Geiger2012CVPR} datasets. Our model ranks second on SceneFlow while being 8 times faster than the best method \cite{cheng2018learning}. Comparing to previous approaches on KITTI, DeepPruner achieves competitive performance to state-of-the-art methods \cite{chang2018pyramid,NIPS2018_7828} and ranks first among all real-time models \cite{mayer2016large,Tonioni_2019_CVPR}. To further showcase the robustness and generalizability of DeepPruner, we evaluate it on the Robust Vision Stereo Challenge \cite{ROB}. Our model achieves state-of-the-art results on multiple datasets \cite{Geiger2012CVPR,schoeps2017cvpr,conf/dagm/ScharsteinHKKNWW14} and obtain first place on the overall ranking.}


\section{Related work}

\paragraph{Classical Stereo Matching:} Estimating disparities (depth) from stereo images has been studied for decades \cite{barnard1982computational}. A stereo algorithm typically consists of the following three steps \cite{scharstein2002taxonomy}: compute a pixel-wise feature representation, construct the cost volume, and final post-processing. As the pixel representation plays a critical role in the process, researchers have exploited a variety of representations, from the simplest RGB values of the surrounding pixels to more discriminative local descriptors such as CENSUS \cite{zabih1994non}, SIFT \cite{lowe2004distinctive}, and BRIEF \cite{calonder2010brief}. Together with the carefully designed post-processing techniques, \eg cost aggregation, semi-global matching \cite{hirschmuller2008stereo}, and Markov random fields \cite{szeliski2008comparative,yamaguchi2014efficient}, they are able to achieve good performance on relatively simple scenarios. 

\paragraph{Deep Stereo Matching:}
In order to further deal with more complex real world scenes, especially texture-less regions  or reflective surfaces, modern approaches leverage CNNs to extract robust features \cite{luo2016efficient} and conduct matching \cite{zagoruyko2015learning,zbontar2016stereo}.  
While these techniques have demonstrated great performance in benchmarks \cite{Geiger2012CVPR}, time-consuming post-processing is still required. With this in mind, researchers propose to directly regress sub-pixel disparities from the given stereo images \cite{yang2018segstereo,cheng2018learning,song2018edgestereo}. By implementing the full traditional stereo pipeline as neural network layers, these models can be trained in an end-to-end fashion and are able to conduct inference completely on GPU, which drastically improves the efficiency. Unfortunately, due to the large size  of the cost volume and aggregation, the memory consumption and the required computation are still very high, rendering the models impractical \cite{kendall2017end,chang2018pyramid}. In this paper, we build our work upon \cite{kendall2017end,chang2018pyramid}. Instead of searching for the full disparity space, we exploit a  novel differentiable version of  PatchMatch to learn to prune out unlikely matches and reduce the complexity of the search space. Our model is able to run in real-time  while maintaining comparable performance. 

\paragraph{PatchMatch:} The seminal work of PatchMatch (PM) was proposed by Barnes \etal in 2009 \cite{barnes2009patchmatch}. It was originally introduced as an efficient way to find dense correspondences  across images for structural editing. The key idea behind it is that, a large number of random samples often lead to good guesses. Additionally, neighboring pixels usually have coherent matches. Therefore, once a good match is found, we can efficiently propagate the information to the neighbors. Due to its effectiveness in pruning out the search space, PatchMatch has drawn wide attention across the community \cite{korman2016coherency,he2012computing, generalizedpm, besse2014pmbp} and has been extended and applied in multiple domains. For instance, Korman and Avidan \cite{korman2016coherency} incorporate the idea of image coherence into Locality Sensitivity Hashing and significantly improve the speed. He and Sun \cite{he2012computing} combine KD-trees with PatchMatch to perform more efficient neighbor matching.  
PatchMatch has also been applied in the stereo setting for fast correspondence estimation \cite{lu2013patch} and slanted plane fitting \cite{bleyer2011patchmatch}. \weichiu{To improve sub-pixel accuracy, Besse \etal \cite{besse2014pmbp} further combines PatchMatch with particle belief propagation and extend it to a continuous MRF inference algorithm. Specifically, \cite{besse2014pmbp}  exploits PatchMatch to overcome the infeasibility of searching over continuous output space. Note that if the  MRF  has only unary terms, it reduces to a k-particle generalized PatchMatch \cite{generalizedpm}.}

Our pruning module is largely inspired by \cite{besse2014pmbp}. We first implement the particle PatchMatch operations as neural network layers and unroll them in a recurrent fashion. We then predict the disparity confidence range as to approximate the marginal distribution of each pixel. Through efficient sampling and propagation, we are able to prune out the solution space effectively and significantly speed up  inference. Importantly, all our operations are differentiable, and thus our approach can be learned end-to-end. 

\paragraph{Real-time Stereo:} Besides our work, there has been several concurrent efforts pushing towards real-time deep learning based stereo estimation \cite{khamis2018stereonet,yin2018hierarchical}. Our work is different from theirs since we adaptively prune out the search space for each region. 
In contrast, they employ a fixed, coarse-to-fine procedure to iteratively find the match. 

\begin{figure}

\includegraphics[width=0.98\linewidth]{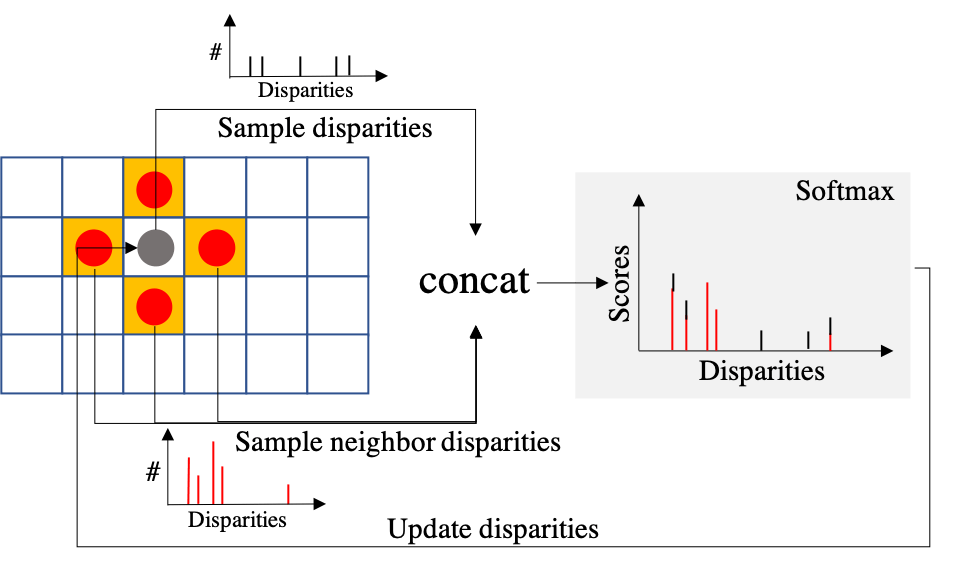}
\caption{\textbf{Illustration of the differentiable patch match operations.}}
\label{fig:patch-match}
\end{figure}

\begin{figure}
\includegraphics[width=0.98\linewidth, trim={5mm 110mm 30mm 5mm},clip]{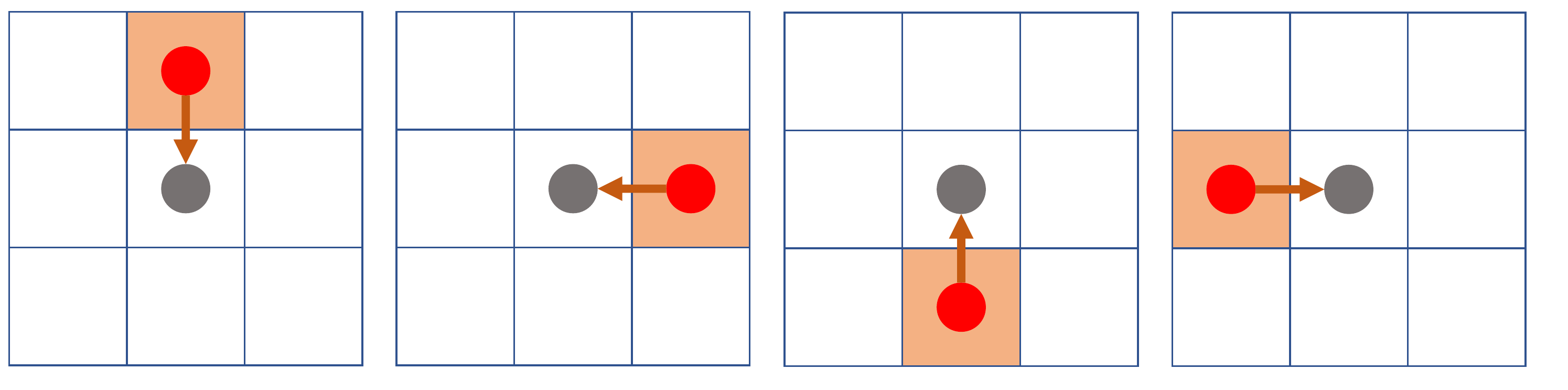}
\caption{\textbf{One hot filter banks within the propagation layer.}}
\label{fig:one-hot}
\end{figure}


\begin{figure}[tb]
\centering
\setlength{\tabcolsep}{1pt}
\begin{tabular}{ccc}
\raisebox{12px}{\rotatebox{90}{RGB}}
\includegraphics[width=0.32\linewidth]{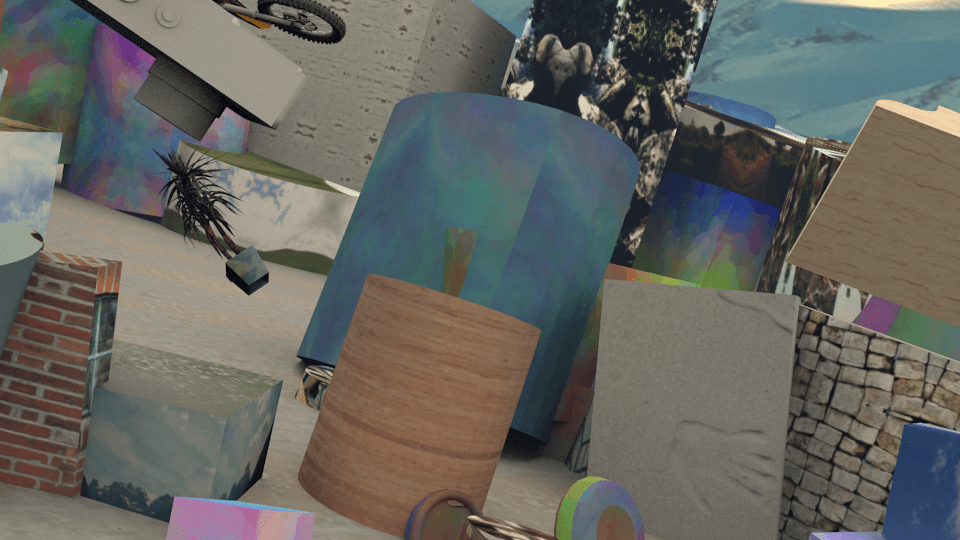}
&\includegraphics[width=0.32\linewidth]{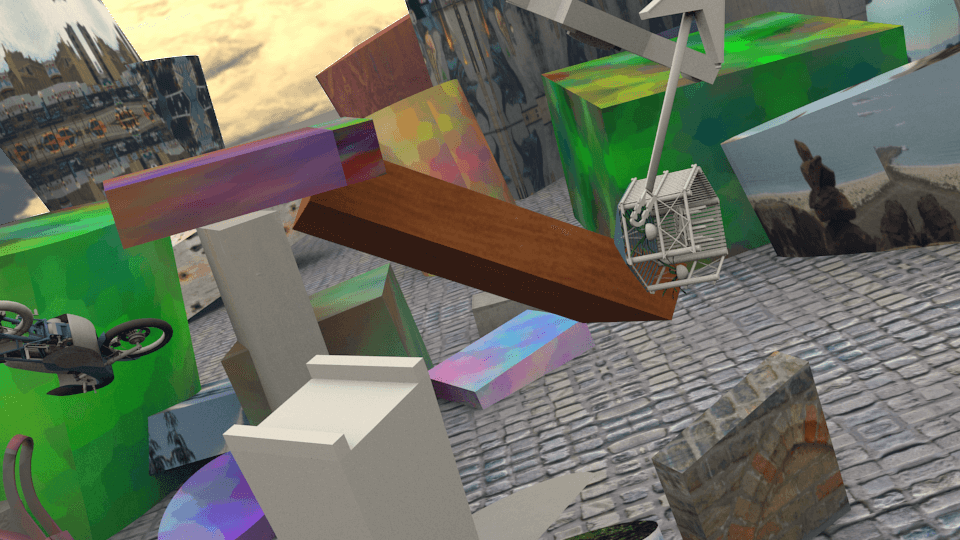}
&\includegraphics[width=0.32\linewidth]{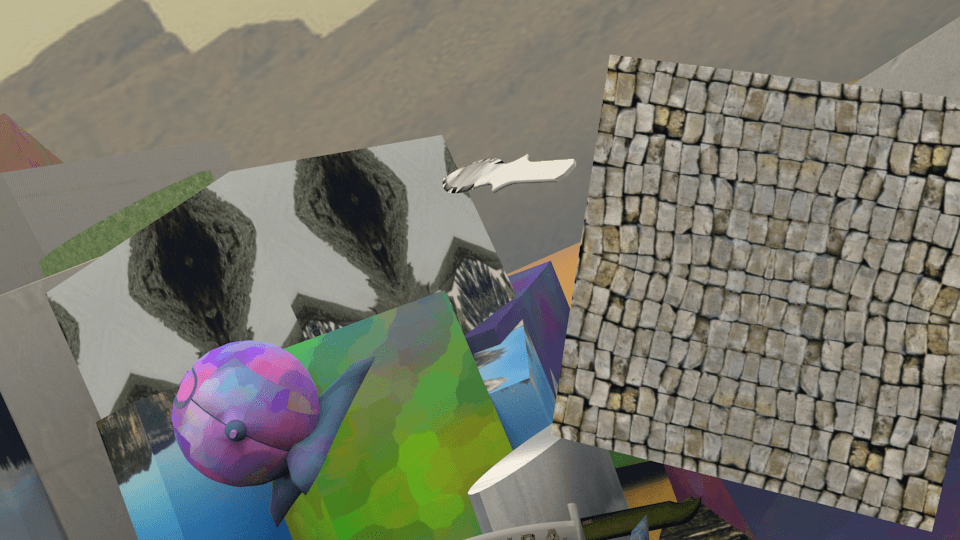}\\
\raisebox{12px}{\rotatebox{90}{GT}}
\includegraphics[width=0.32\linewidth]{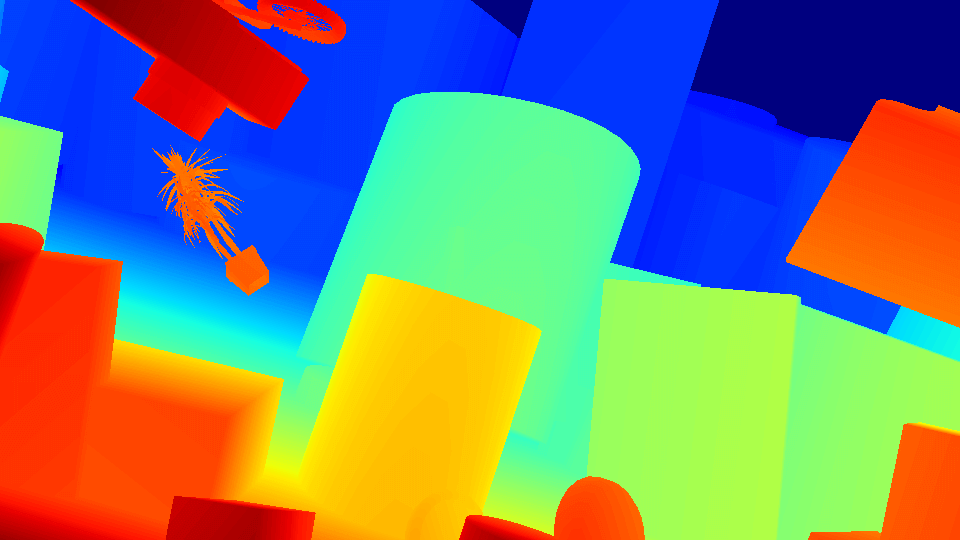}
&\includegraphics[width=0.32\linewidth]{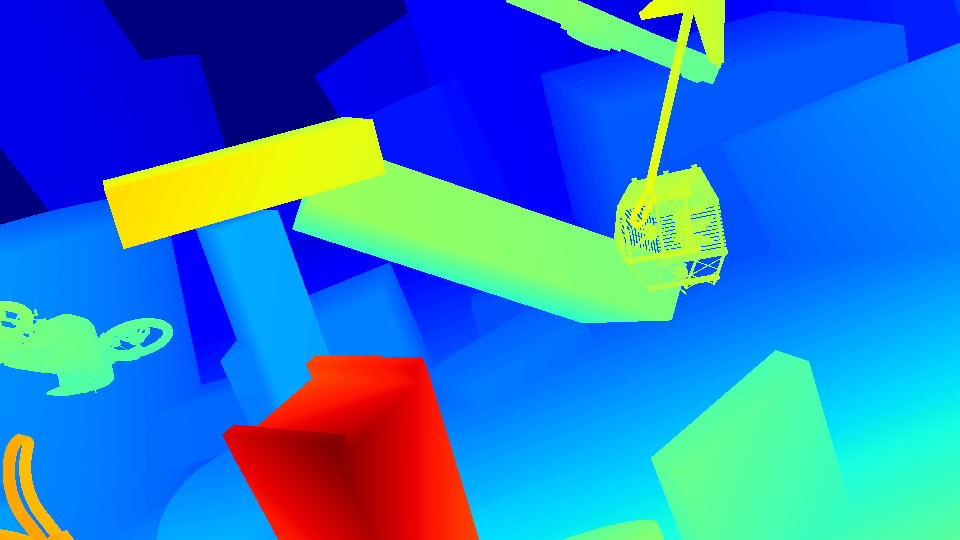}
&\includegraphics[width=0.32\linewidth]{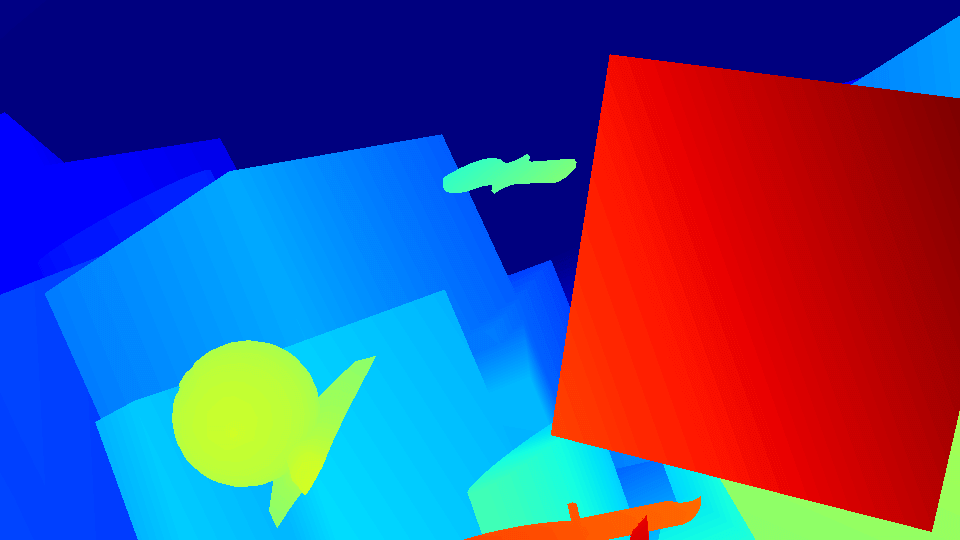}\\
\raisebox{2px}{\rotatebox{90}{Our-Best}}
\includegraphics[width=0.32\linewidth]{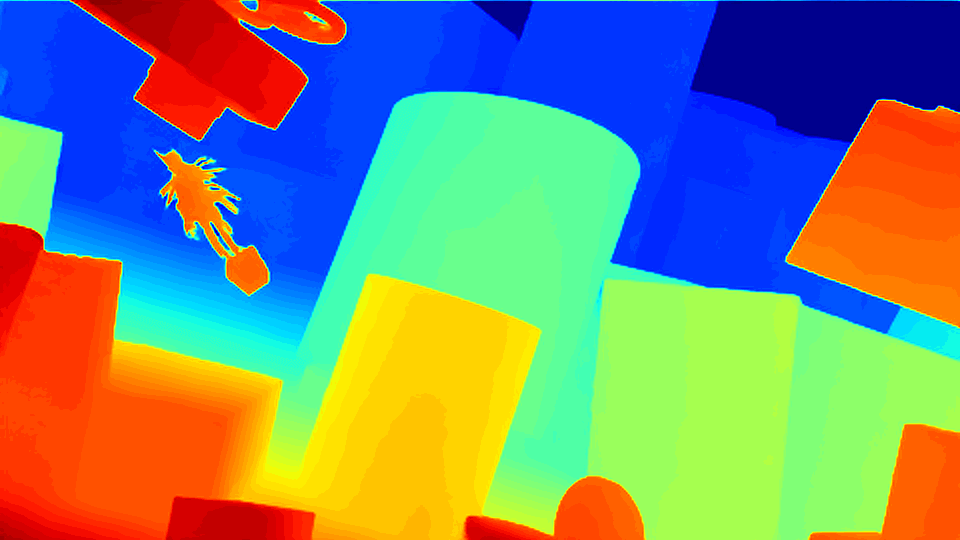}
&\includegraphics[width=0.32\linewidth]{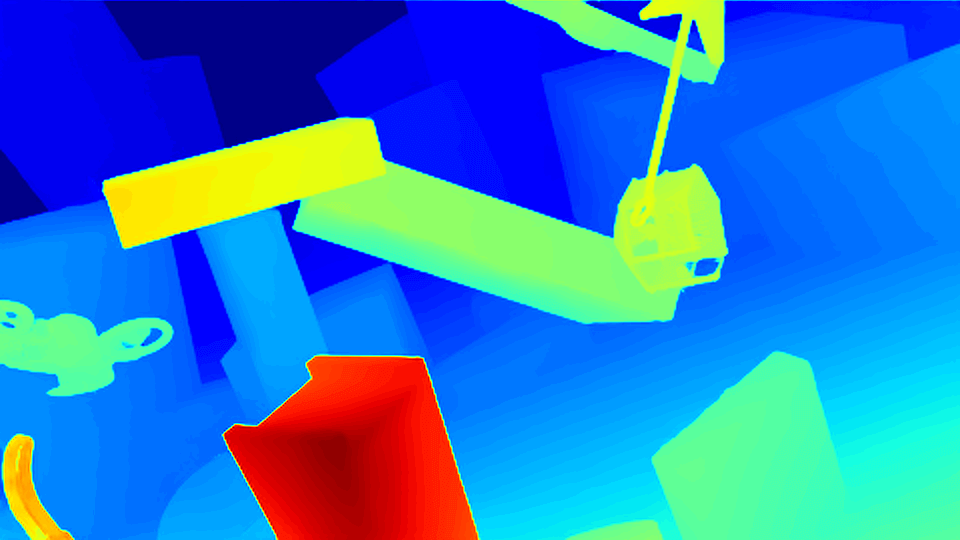}
&\includegraphics[width=0.32\linewidth]{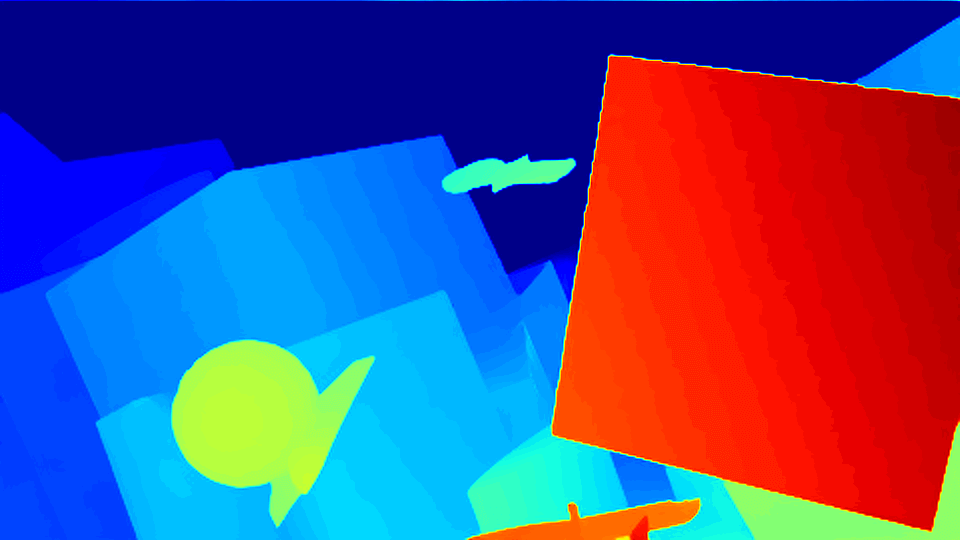}
\\\end{tabular}
\caption{\textbf{Qualitative results on SceneFlow dataset.} }
\label{fig:qual-scene}
\end{figure}

\section{Learning to Prune for Stereo Matching}

Our aim is to design an efficient stereo algorithm that not only produces reliable and accurate estimations, but also runs in real-time.
Towards this goal, we present a simple yet effective solution that combines deep learning with PatchMatch to prune out the potentially large search space and significantly speeds up  inference.

We start our discussion by describing the feature extraction backbone. We then briefly review the PatchMatch algorithm and show that it can be naturally incorporated into neural networks to prune out the search space and speed up cost volume construction. Finally, we describe how to aggregate the cost, refine the estimation, and perform end-to-end learning. We refer the reader to  Fig. \ref{fig:overview} for an illustration of our approach.

\subsection{Feature Extraction}
\label{sec:feature}
The goal of the feature extraction network is to produce a reliable pixel-wise feature representation from the input image. More formally, given a pair of stereo images $\{\mathbf{x}_0, \mathbf{x}_1\}$, we seek to learn a set of deep features $\mathbf{f}_0, \mathbf{f}_1$ that is useful for matching.
Towards this goal, following \cite{kendall2017end,chang2018pyramid}, we exploit a 2D convolutional neural network with a spatial pyramid pooling module \cite{he2014spatial,zhao2017pyramid} as our backbone. Specifically, we employ four residual blocks and use $\times2$ dilated convolution for the last block to enlarge the receptive field.  We then apply spatial pyramid pooling to build a 4-level pyramid feature. Through multi-scale information, the model is able to capture large context while maintaining a high spatial resolution. 
The size of the final feature map is 1/4 of the original input image size.
We share the parameters for the left and right feature network.
Now that we have a reliable feature representation for each pixel, the next step is to construct the cost volume.

\subsection{Pruning Through Differentiable PatchMatch}
Modern stereo approaches typically generate a cost volume over the full disparity space \cite{chang2018pyramid,kendall2017end,cheng2018learning,song2018edgestereo}. The large search space not only increases memory consumption, but also intensifies the computational burden. For example, considering PSM-Net \cite{chang2018pyramid},  3D cost volume construction and aggregation takes more than 250 ms. These two operations themselves render the real-time applications infeasible. 
In this paper, we tackle this issue by designing an efficient PatchMatch-based pruner module that is able to predict a confidence range for each pixel, and construct a sparse cost volume that requires significantly less operations.
This allows the model to focus only on high likelihood regions and save large amount of computation and memory. Unlike standard PatchMatch, our module is differentiable, making end-to-end learning possible. 
Importantly, as shown in our experiments, such confidence range is also a promising indicator of uncertainty and a foreseer of potential prediction error. This is important when depth estimates are used in downstream tasks.

\paragraph{PatchMatch revisited:}
Our pruner module is motivated by the elegant and classic PatchMatch algorithm \cite{barnes2009patchmatch}. PatchMatch methods \shivam{ \cite{bleyer2011patchmatch,besse2014pmbp,barnes2009patchmatch,generalizedpm} }  typically consist of the following three steps\footnote{For simplicity, we omit the local random resampling applied to the current particles.}:
\begin{enumerate}
	\item \textbf{Particle sampling}: generate $k$ random candidates;
	\item \textbf{Propagation}: propagate particles to neighbors;
	\item \textbf{Evaluation}: update best $k$ solutions by evaluating current and propagated particles; 
\end{enumerate}
Once the initialization is done (step 1,), the greedy approach iterates between step 2 and step 3 until convergence or a fix number of steps is reached.
In practice, this often leads to good results without enumerating a,ll possibilities. Originally, $k$ was set to 1 \cite{barnes2009patchmatch}. But later on,  generalized PatchMatch  \cite{generalizedpm} draw connections to particle sampling methods and extend PatchMatch to utilize the top-$k$. This not only increases the expressive power but also enables faster convergence.

\paragraph{Differentiable PatchMatch:}  

In this work, we unroll generalized PatchMatch as a recurrent neural network, where each unrolling step is equivalent to each iteration of the algorithm. 
This is important as it  allow us to train our full model end-to-end. 
Specifically, we design the following layers: 
\begin{enumerate}
	\item \textbf{Particle sampling layer}: for each pixel $i$, we randomly generate $k$ disparity values from the uniform distribution over predicted/pre-defined search space;
	\item \textbf{Propagation layer}: particles from adjacent pixels are propagated together through convolution with a pre-defined one-hot filter pattern (see Fig. \ref{fig:one-hot}), which encodes the fact that  we allow each pixel to propagate particles to its 4-neighbours. 
	\item \textbf{Evaluation layer}: for each pixel $i$, matching scores are computed by taking the inner product between the left feature and the right feature: $s_{i, j} = \langle \mathbf{f}_0(i), \mathbf{f}_1(i + d_{i, j}) \rangle$ for all candidates $j$. The best $k$ disparity value for each pixel is carried towards the next iteration.
\end{enumerate}
Our architecture design has   one particle sampling layer at the bottom, and then iterates through propagation and evaluation layers recurrently. 
As the $\arg\max$ operator during evaluation is not differentiable, we replace it with a soft version \cite{kendall2017end}:
\begin{equation}
\label{eq:softargmaxs}
\hat{d_i} = \frac{\sum_j s_{i, j} \cdot d_{i, j}}{\sum_j s_{i, j}}.
\end{equation}
Fig.~\ref{fig:patch-match} depicts the computation graph of one recurrent step, which combines propagation and evaluation.  
In practice, rather than allowing  each particle to reside in the full disparity space, we divide the search space into $k$ intervals, and force  the $i-$th particle to be in a $i-$th interval. 
This guarantees the diversity of the particles and helps improve accuracy for later computations, which we show  in our experiments. 
Since all operations are differentiable, we can directly back-propagate through all unrolling steps and train the model in an end-to-end fashion.

\begin{figure*}[tb]
\centering
\setlength{\tabcolsep}{1pt}
\def\arraystretch{0.7}
\begin{tabular}{ccccc}
\raisebox{8px}{\rotatebox{90}{RGB}}
\includegraphics[width=0.24\linewidth]{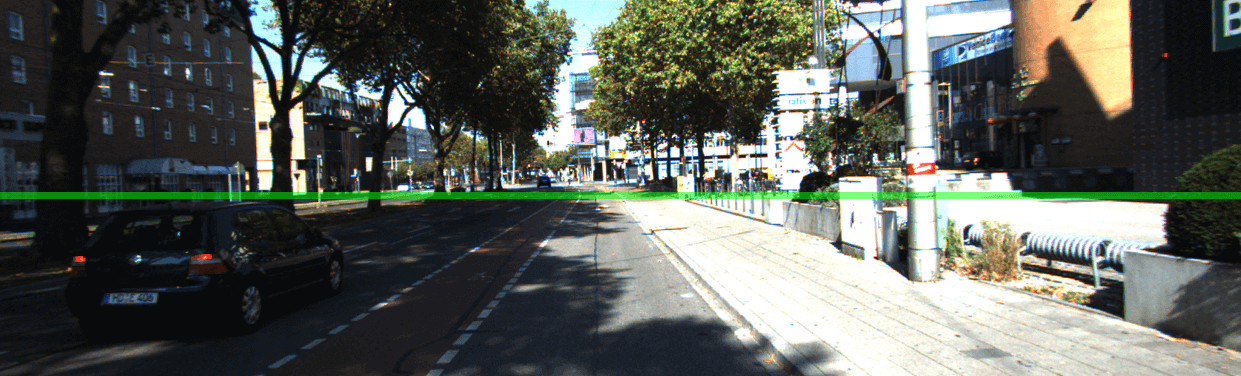}
&\includegraphics[width=0.24\linewidth]{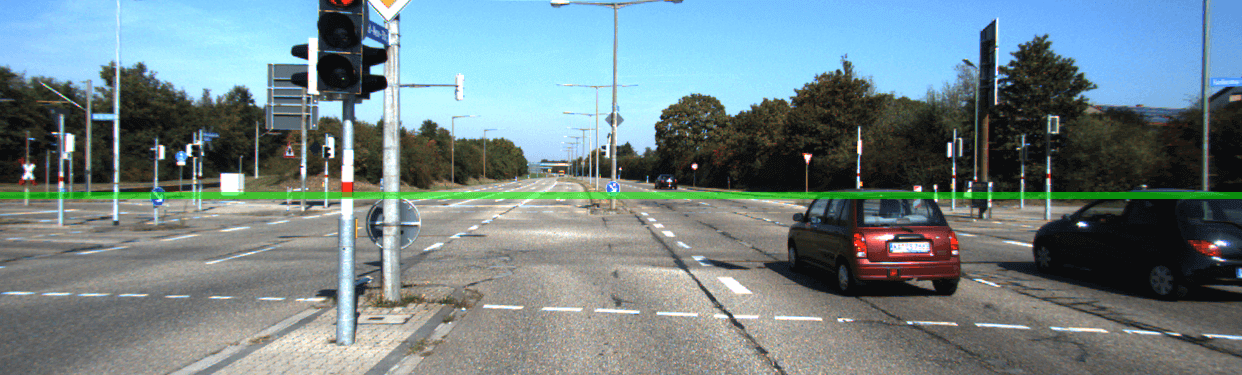}
&\includegraphics[width=0.24\linewidth]{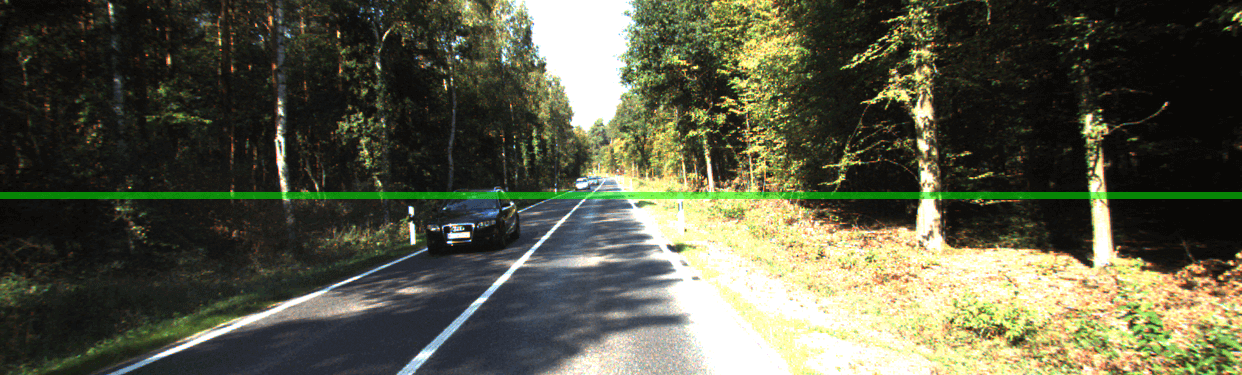}
&\includegraphics[width=0.24\linewidth]{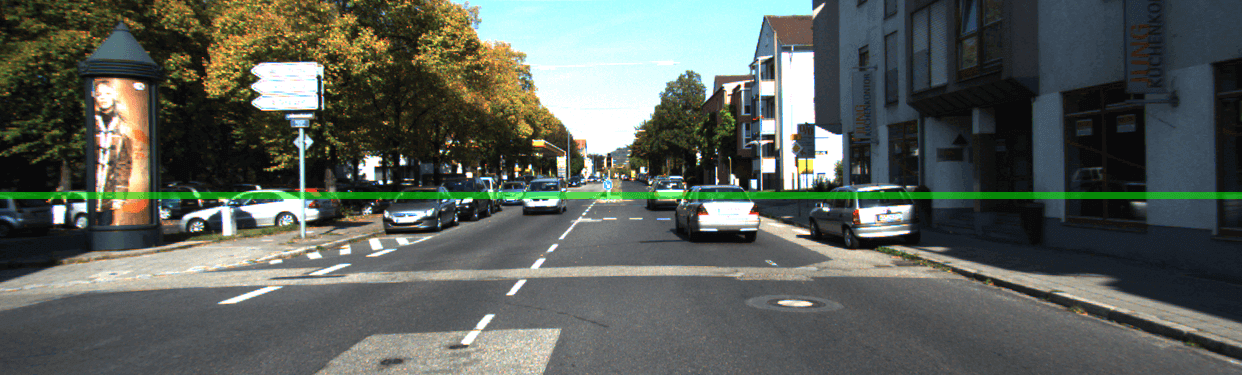}\\
\raisebox{12px}{\rotatebox{90}{GT}}
\includegraphics[width=0.24\linewidth]{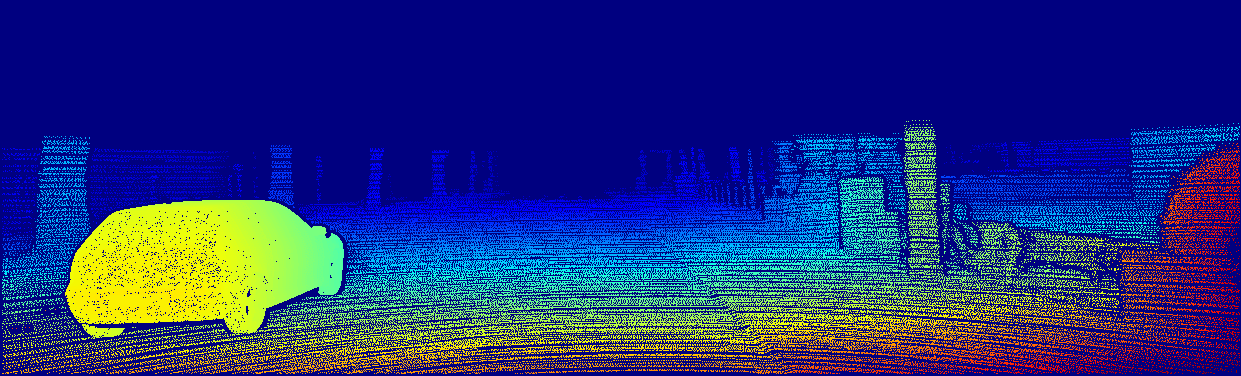}
&\includegraphics[width=0.24\linewidth]{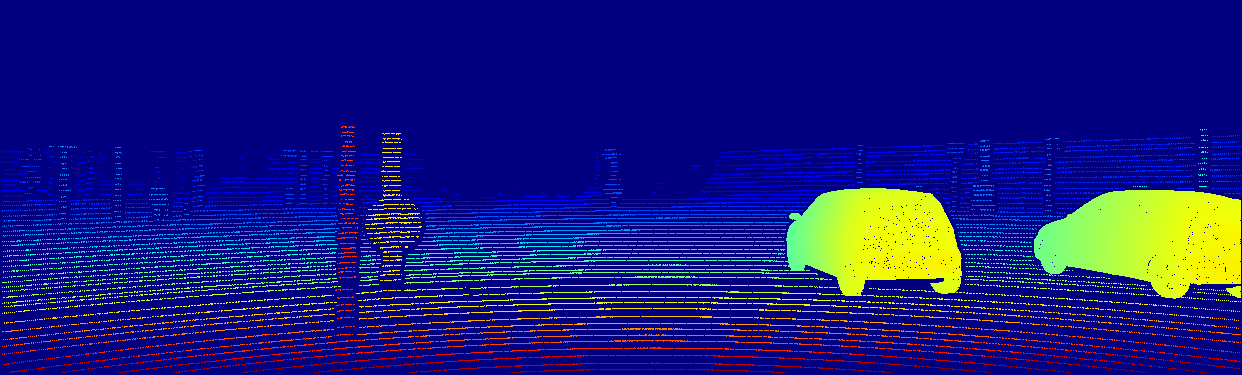}
&\includegraphics[width=0.24\linewidth]{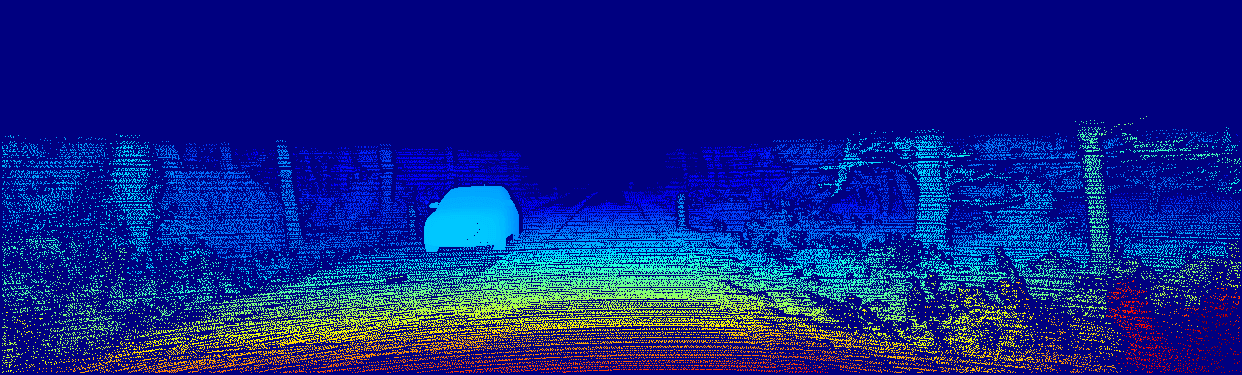}
&\includegraphics[width=0.24\linewidth]{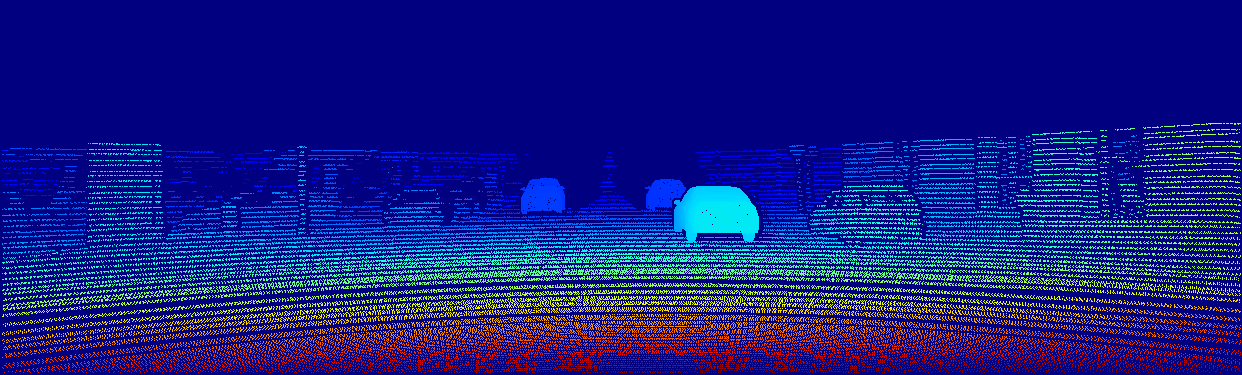}\\
\raisebox{0px}{\rotatebox{90}{Our-Best}}
\includegraphics[width=0.24\linewidth]{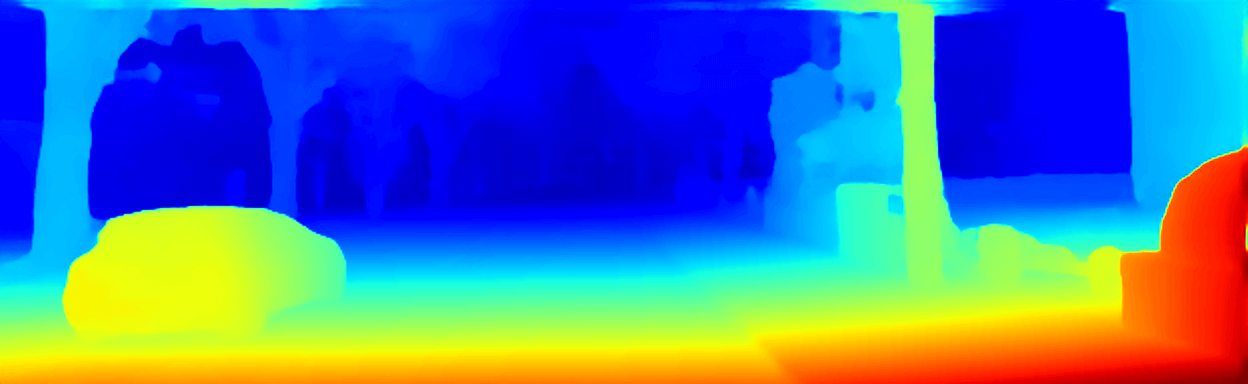}
&\includegraphics[width=0.24\linewidth]{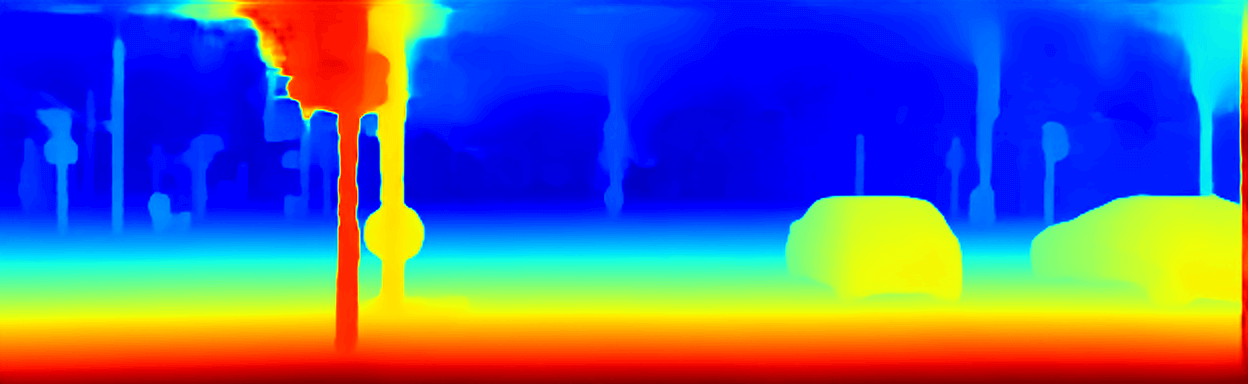}
&\includegraphics[width=0.24\linewidth]{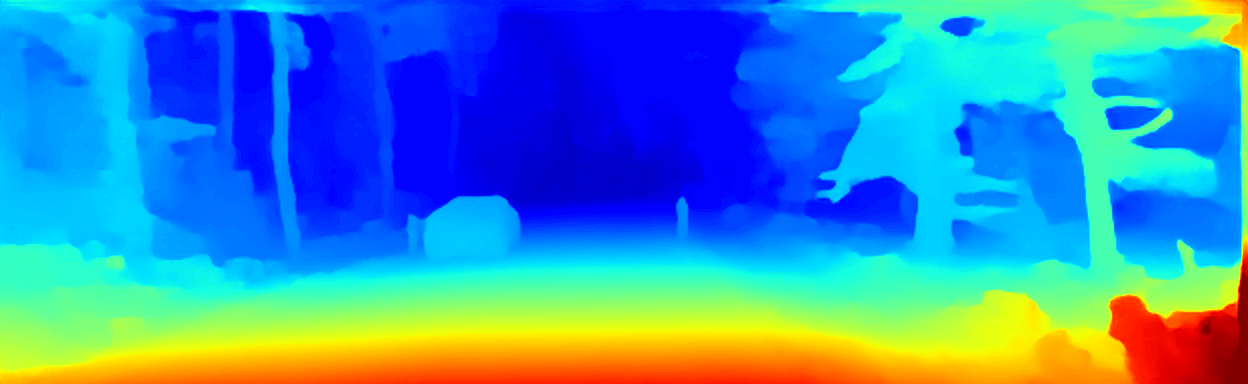}
&\includegraphics[width=0.24\linewidth]{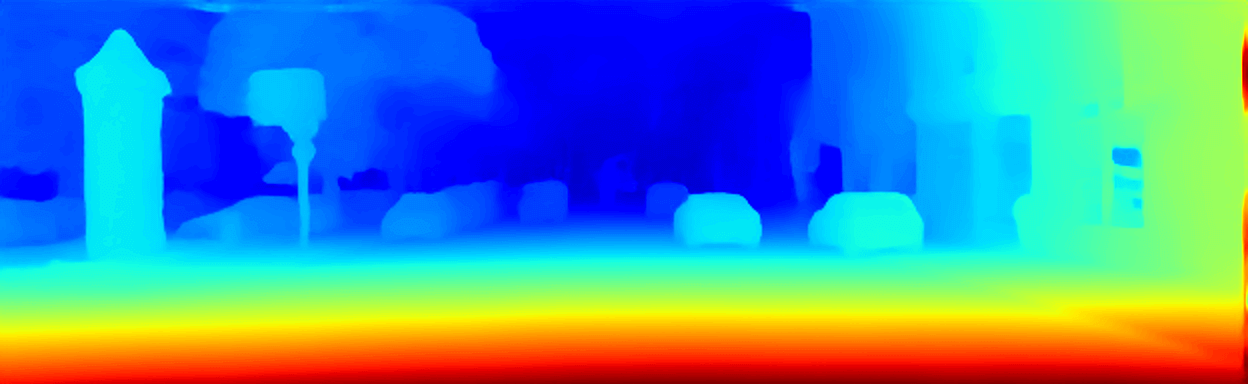}\\
\raisebox{8px}{\rotatebox{90}{Confidence}}
\includegraphics[trim={1.2cm 0.5cm 1.5cm 1.2cm},clip, width=0.24\linewidth]{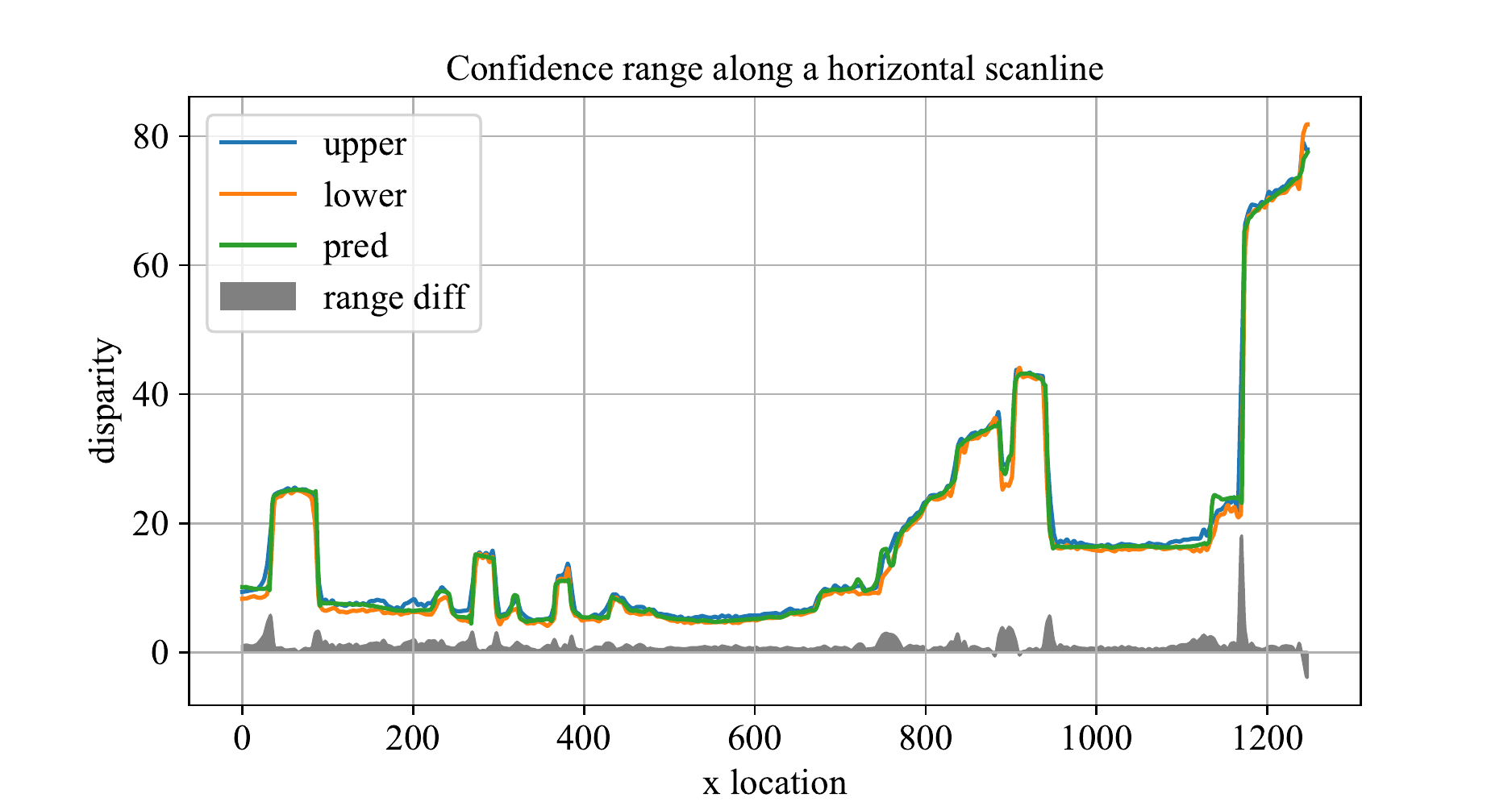}
&\includegraphics[trim={1.2cm 0.5cm 1.5cm 1.2cm},clip, width=0.24\linewidth]{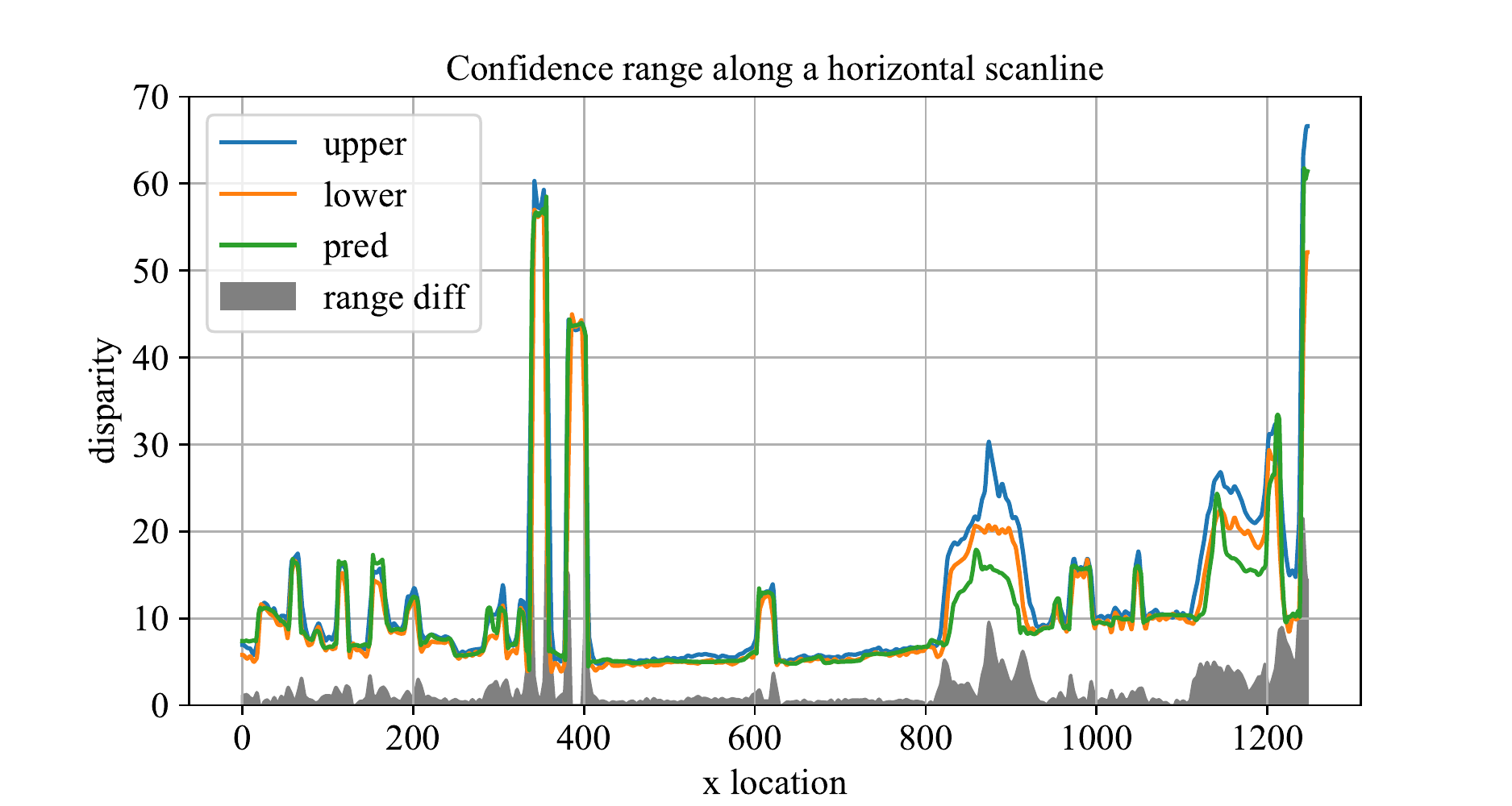}
&\includegraphics[trim={1.2cm 0.5cm 1.5cm 1.2cm},clip, width=0.24\linewidth]{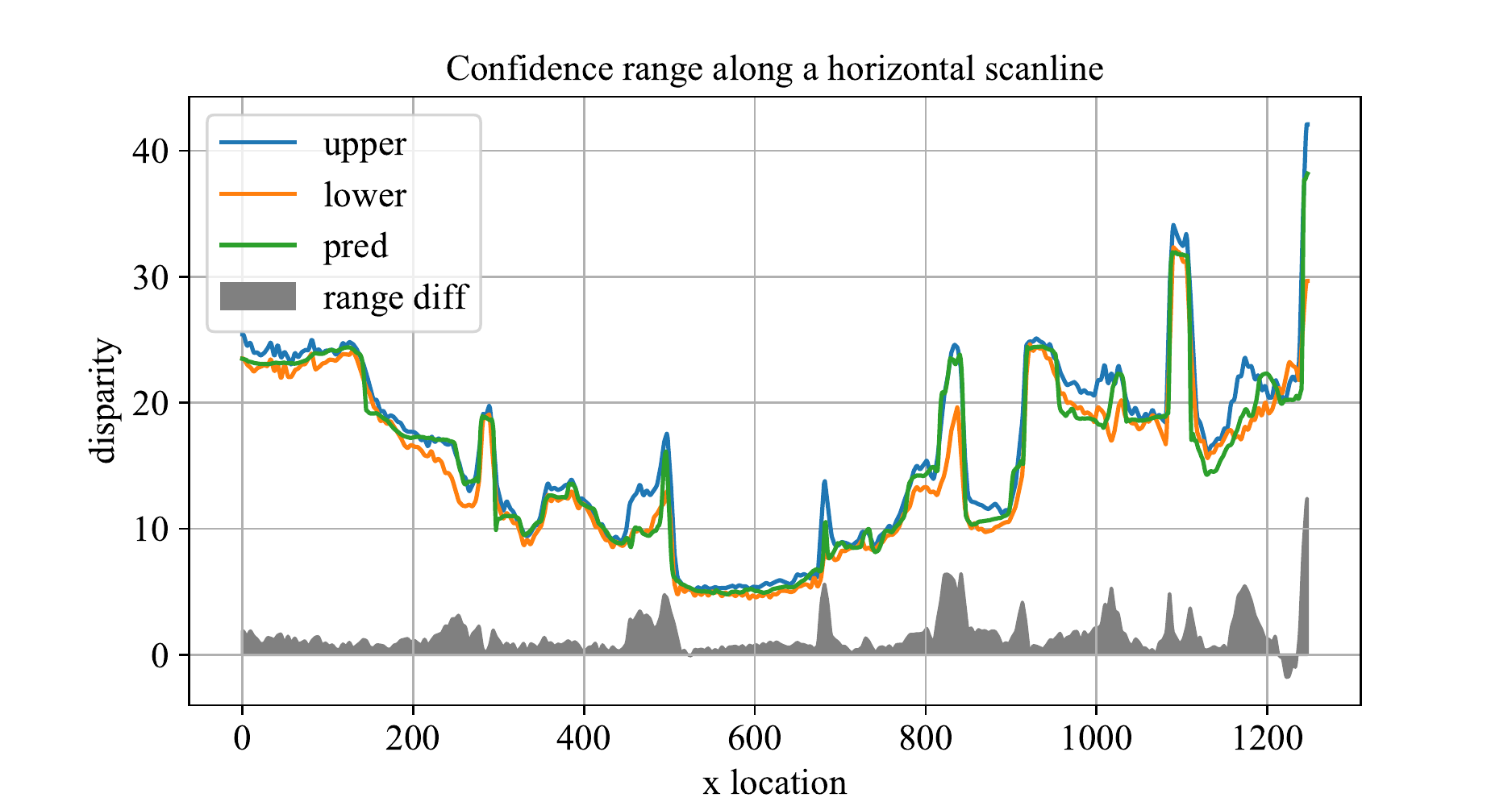}
&\includegraphics[trim={1.2cm 0.5cm 1.5cm 1.2cm},clip, width=0.24\linewidth]{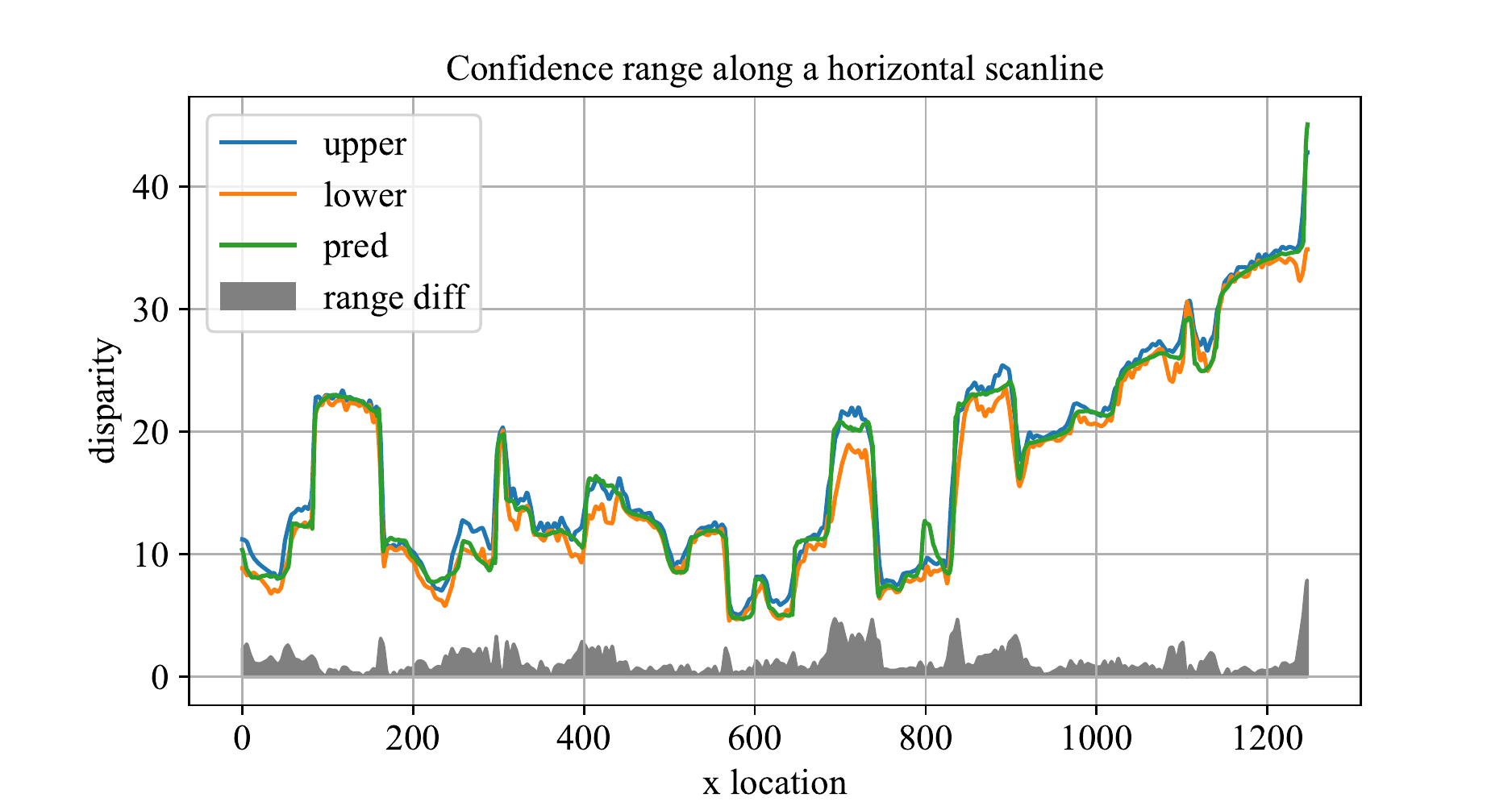}\\\end{tabular}
\caption{\textbf{Visualization of confidence range predictor.} \weichiu{At the bottom row, we show the predicted search range for pixels along the horizontal green line on validation images shown at the top row. \slwang{Blue and orange line represent the upper and lower bound of the search range, respectively, and the grey shades depict the range size.}
For most pixels, DeepPruner predicts a very small search range, thereby allowing efficient cost aggregation. Higher search ranges often happen at boundary pixels, or pixels that are occluded in one view.}
}
\label{fig:qual-uncertainty}
\end{figure*}
\paragraph{Confidence range prediction:}\label{CRP} The original search space for all pixels is identical.  
However, in practice, for each pixel, the highly probable disparities lie in  a narrow region. Using the small subset of disparities estimated from the PatchMatch stage, we have sufficient information to predict the range in which the true disparity lies. We thus exploit a confidence range prediction network to adjust the search space for each pixel.
The network has a convolutional encoder-decoder structure. It takes the sparse disparity estimations from the differentiable PatchMatch, the left image and the warped right image (warped according to the sparse disparity estimations) as input and outputs
a confidence range $\mathcal{R}_i = [l_i, u_i]$ for each pixel $i$.  

The confidence range prunes out the space of unlikely matches, allowing the expensive cost-volume construction and aggregation to happen  only at a few disparity values. 

\subsection{Cost Aggregation and Refinement}
\label{sec:cost}

\paragraph{Cost aggregation: } Based on the predicted range in the pruning module, we build the 3D cost volume estimator and conduct spatial aggregation. Following  common practice \cite{chang2018pyramid,kendall2017end},  we take the left image, the warped right image and corresponding disparities as input and output the cost over the disparity range at the size $B\times R \times H \times W$, where $R$ is the number of disparities per pixel. Compared to prior work \cite{chang2018pyramid,kendall2017end}, our $R$ is more than 10 times smaller, making this module very efficient. Soft-$\arg\max$ defined in Eq.~\ref{eq:softargmaxs} is again used to predict the disparity value $\mathbf{y}_\textrm{cost}$, so that our approach is end-to-end trainable. 

\paragraph{Refinement:}

We utilize a lightweight fully convolutional refinement network to further boost the performance. The network takes left image convolutional features from the 
second residual block of the feature network and the current disparity estimation $\mathbf{y}_\textrm{cost}$ as input. It then outputs the finetuned disparity prediction $\mathbf{y}_\textrm{refine}$. The low-level feature information serves as a guidance to reduce noise and improve the quality of the final disparity map, especially on sharp boundaries.

\begin{table*}[tb]
\centering
\scalebox{0.8}{
\begin{tabular}{lccccccc|cc}
\specialrule{.2em}{.1em}{.1em}

&GC-Net \cite{kendall2017end}& SegStereo \cite{yang2018segstereo}& CRL \cite{pang2017cascade}& PDS-Net \cite{NIPS2018_7828}& PSM-Net \cite{chang2018pyramid}&CSPN \cite{cheng2018learning}&\underline{\smash{Our-Best}}  &DispNetC \cite{mayer2016large} & \underline{\smash{Our-Fast}}\\
\hline
EPE&2.51&1.45&1.32&1.12&1.09&\textbf{0.78}&0.86 &1.68 & \textbf{0.97} \\
Runtime&900 ms&600 ms&470 ms&500 ms&410 ms&500 ms& \textbf{182 ms} &\textbf{60 ms} & 62 ms \\
\specialrule{.1em}{.05em}{.05em}
\end{tabular}
}
\caption{\textbf{Quantitative results on SceneFlow Dataset.} \weichiu{Our approach ranks $2^{nd}$ among all competing algorithms. Our fast model is 8 times faster than prior art and improves the performance of previous real-time model by $40\%$.}}

\label{tab:quant-scenflow}
\end{table*}

\subsection{End-to-end Learning}
\label{sec:train}
Our network is end-to-end differentiable. We use back-propagation to learn the parameters. Given the GT disparity $\mathbf{y}$, the total loss function is defined as follows:

\[
\ell_{s}(\mathbf{y}_\mathrm{cost}- \mathbf{y}_\mathrm{gt}) + \ell_{s}(\mathbf{y}_\mathrm{refine}- \mathbf{y}_\mathrm{gt}) \] 
\[ + \gamma\{  \ell_{\textrm{lower}}(\mathbf{l}-\mathbf{y}_\mathrm{gt}) + \ell_\textrm{upper}(\mathbf{u}-\mathbf{y}_\mathrm{gt}) \}
\]
where the standard smooth-$\ell_1$ loss is applied over the disparity prediction in cost aggregation stage and final refinement stage respectively. Thus we define:
 \[\ell_\textrm{s}(x) = \left\{ 
\begin{array}{cc}  0.5x^2 & \text{\ if \ } |x| < 1 \\
|x| - 0.5 & \text{otherwise} \\
\end{array} \right.\]
This loss has the advantage of being differentiable everywhere like the $\ell_2$ loss, but more robust to outlier like the $\ell_1$ loss. 
Moreover, the loss over upper and lower bound of the range is defined as a boomerang shape unbalanced smooth-$\ell_1$ loss: 
\[\ell_\textrm{lower}(x) = \left\{ 
\begin{array}{cc}  (1- \lambda) \ell_\textrm{s}(x) & \text{\ if \ } x > 0 \\
\lambda \ell_\textrm{s}(x) &  \text{otherwise} \\
\end{array} \right.\]
 \[\ell_\textrm{upper}(x) = \left\{ 
\begin{array}{cc}  \lambda\ell_\textrm{s}(x) & \text{\ if \ } x > 0 \\
(1- \lambda)\ell_\textrm{s}(x) &  \text{otherwise} \\
\end{array} \right.\]
with $0 < \lambda<0.5$. $\gamma$ is the balancing scalar. 

Note that $\ell_\textrm{upper}$ encourages the upper-bound prediction to be closer to but preferably larger than GT disparity; whereas the $\ell_\textrm{lower}$ pushes the lower-bound prediction to be closer to but preferably smaller than GT disparity.


\begin{table}[tb]
\centering
\scalebox{0.7}{
\begin{tabular}{lcccccccc}
\specialrule{.2em}{.1em}{.1em}

&Inference&\multicolumn{3}{c}{Noc (\%)} &\multicolumn{3}{c}{All (\%)}\\
Methods & Runtime &\emph{bg} &\emph{fg} &\emph{all} &\emph{bg} &\emph{fg} &\emph{all} \\
\hline
Content-CNN \cite{luo2016efficient} & 1000 ms & 3.32 & 7.44 & 4.00 & 3.73 & 8.58 & 4.54 \\
MC-CNN \cite{zbontar2016stereo} & 67000 ms & 2.48 & 7.64 & 3.33 & 2.89 & 8.88 & 3.89 \\
GC-Net \cite{kendall2017end} & 900 ms & 2.02 & 3.12 & 2.45 & 2.21 & 6.16 & 2.87 \\
CRL \cite{pang2017cascade} & 470 ms & 2.32 & 3.68 & 2.36 & 2.48 & 3.59 & 2.67 \\ 
PDS-Net \cite{NIPS2018_7828} & 500 ms & 2.09 & 3.68 & 2.36 & 2.29 & 4.05 & 2.58 \\
PSM-Net \cite{chang2018pyramid} & {410 ms} & 1.71 & 4.31 & 2.14 &1.86 &4.62 &2.32 \\
SegStereo \cite{yang2018segstereo} &600 ms & 1.76 & 3.70 & 2.08 & 1.88& 4.07 & 2.25 \\
EdgeStereo \cite{song2018edgestereo} &700 ms & 1.72 & 3.41 & 2.00 & 1.87 & 3.61 & 2.16 \\
CSPN \cite{cheng2018learning} &500 ms & \textbf{1.40} & \textbf{2.67} & \textbf{1.61} & \textbf{1.51} & \textbf{2.88} & \textbf{1.74} \\
{\href{http://www.cvlibs.net/datasets/kitti/eval_scene_flow_detail.php?benchmark=stereo&result=c8972e144fae42dd1071091b39e5437f43753269}{\underline{\smash{DeepPruner-Best}}}} &\textbf{182 ms} & 1.71 & 3.18 & 1.95 & 1.87 & 3.56 & 2.15 \\
\hline
MAD-Net \cite{Tonioni_2019_CVPR} & \textbf{20 ms} & 3.45 & 8.41 & 4.27 & 3.75 & 9.2 & 4.66\\ 
DipsNetC \cite{mayer2016large} & 60 ms & 4.11 & 3.72 & 4.05 & 4.32 & 4.41 & 4.34 \\
{\href{http://www.cvlibs.net/datasets/kitti/eval_scene_flow_detail.php?benchmark=stereo&result=fde5e04e47f4c5b8c40ec3584eff4d6001b2a7fb}{\underline{\smash{DeepPruner-Fast}}}} & 61 ms & \textbf{2.13} & \textbf{3.43} & \textbf{2.35} & \textbf{2.32} & \textbf{3.91} & \textbf{2.59} \\

\specialrule{.1em}{.05em}{.05em}
\end{tabular}
}
\caption{\textbf{Quantitative Results on KITTI 2015 Test Set.}  
\weichiu{\textbf{Top:} Our model acheives comparable performance to state-of-the-art models while being significantly faster. \textbf{Bottom:} Comparing to other real-time methods (\emph{e.g.} DispNet), our stereo estimation is much more precise.}}
\label{tab:quant-2015}
\end{table}

\begin{table*}[tb]
\centering
\scalebox{0.9}{
\begin{tabular}{lcccccc}
\specialrule{.2em}{.1em}{.1em}

Module & Feature Extraction & PatchMatch-1 & Confident Range & PatchMatch-2 & Cost Aggregation & RefineNet\\ \hline
DeepPruner-Best & 54 ms & 20 ms& 61 ms & 13 ms & 32 ms & 3 ms\\
DeepPruner-Fast & \cellcolor{white!25}28 ms & \cellcolor{white!25}5 ms& \cellcolor{white!25}16 ms &\cellcolor{white!25} 3 ms &\cellcolor{white!25} 8 ms & \cellcolor{white!25}4 ms\\
\specialrule{.1em}{.05em}{.05em}
\end{tabular}
}
\caption{\textbf{Runtime Breakdown:} 
\weichiu{As PatchMatch and Confidence Range Predictor significantly reduce the search space, we only conduct cost aggregation on a small subset of disparities, thereby being much faster.}}
\label{tab:runtime}
\end{table*}

\begin{figure*}[tb]
\centering
\def\arraystretch{0.7}
\setlength{\tabcolsep}{1pt}
\begin{tabular}{ccccc}
\raisebox{8px}{\rotatebox{90}{\small RGB}}
\includegraphics[width=0.24\linewidth]{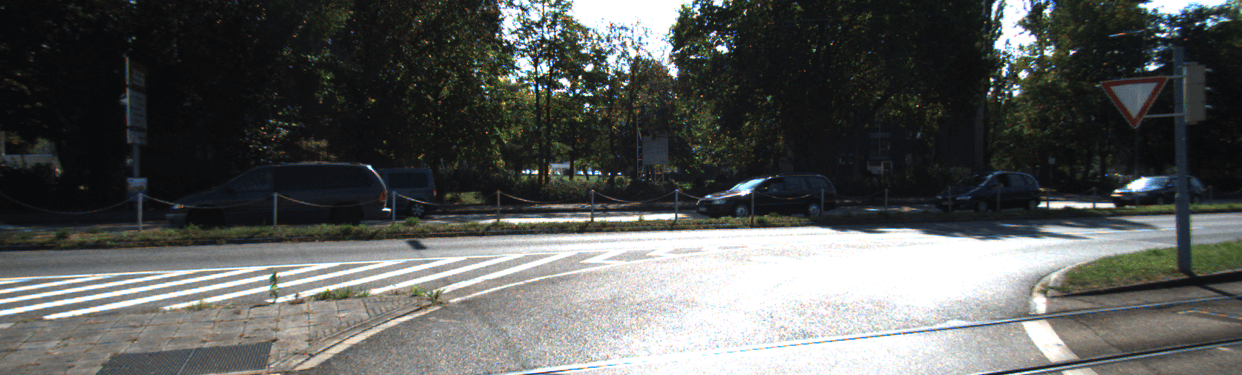}
&\includegraphics[width=0.24\linewidth]{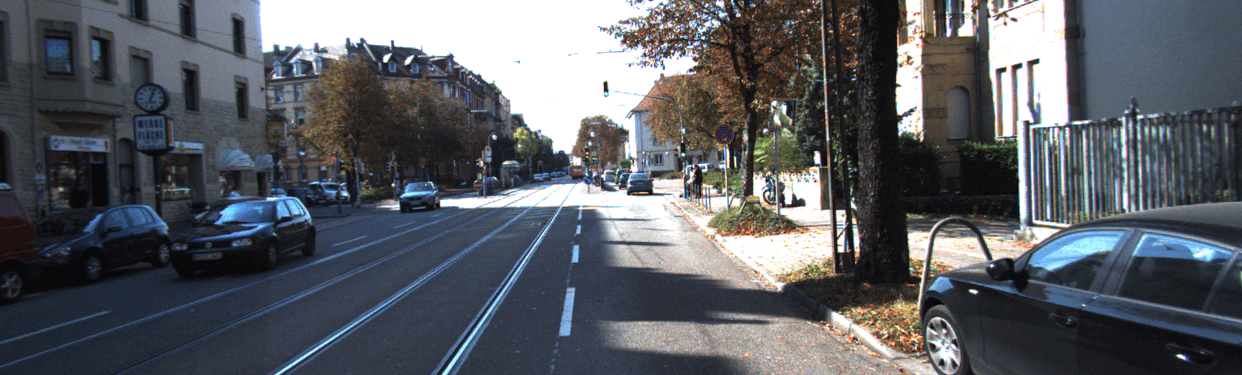}
&\includegraphics[width=0.24\linewidth]{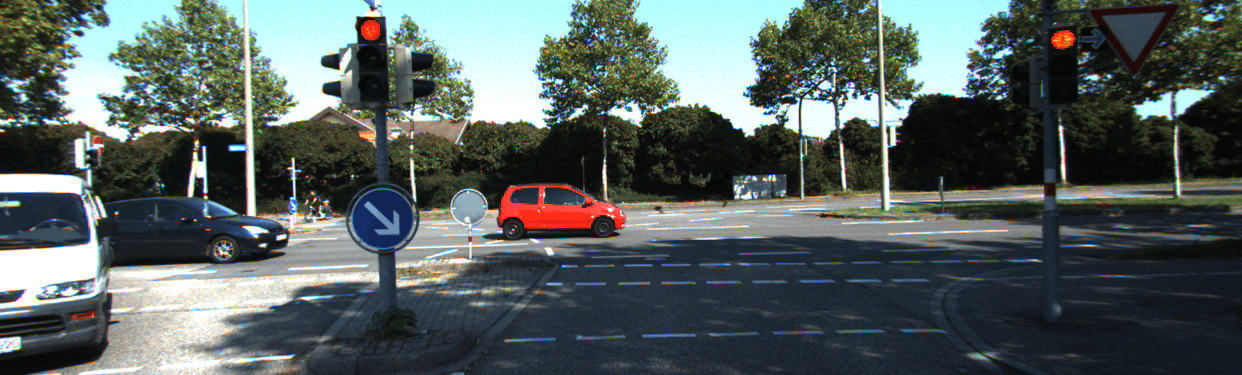}
&\includegraphics[width=0.24\linewidth]{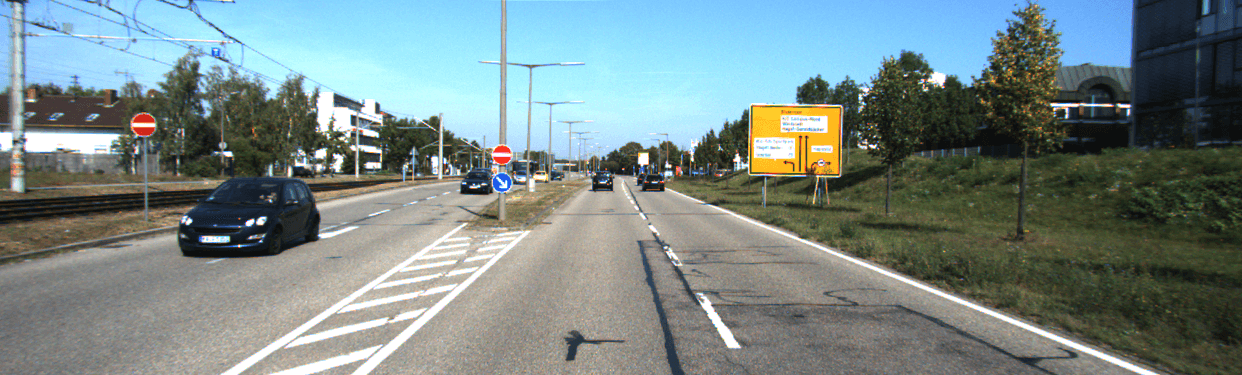}\\
\raisebox{8px}{\rotatebox{90}{\small CSPN}}
\includegraphics[width=0.24\linewidth]{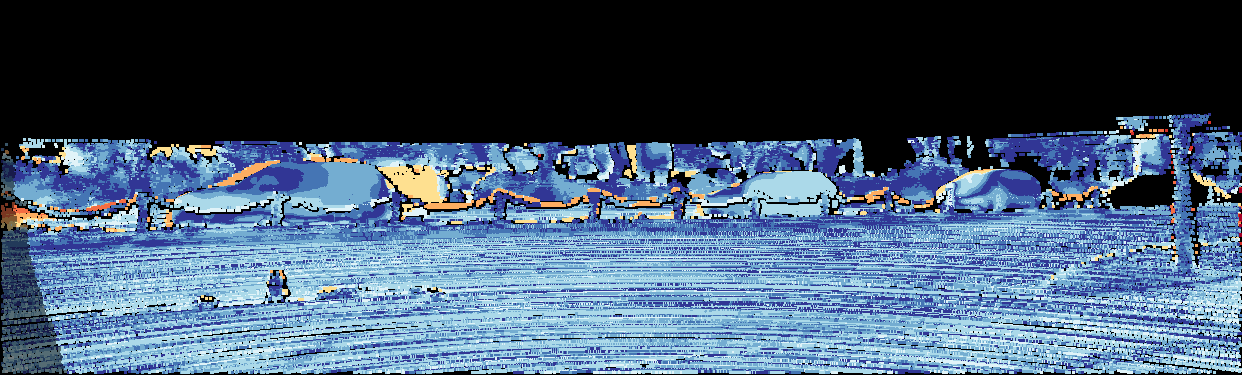}
&\includegraphics[width=0.24\linewidth]{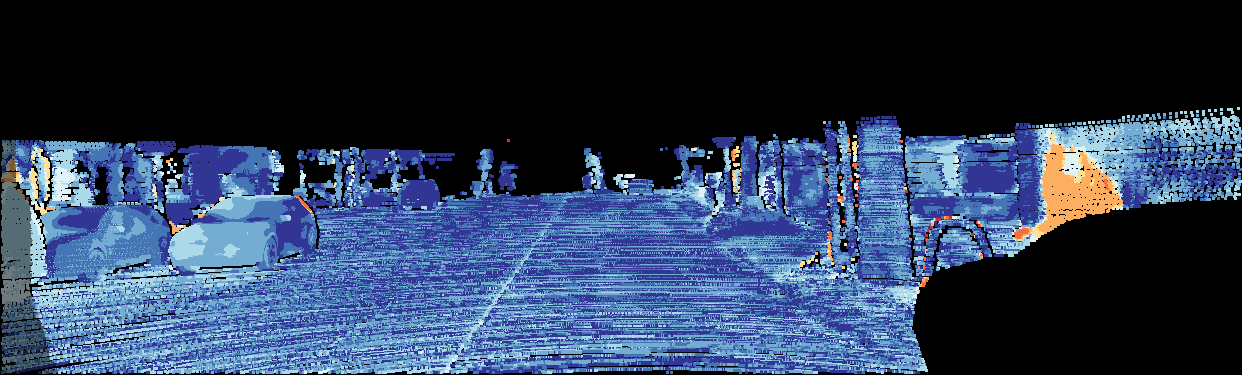}
&\includegraphics[width=0.24\linewidth]{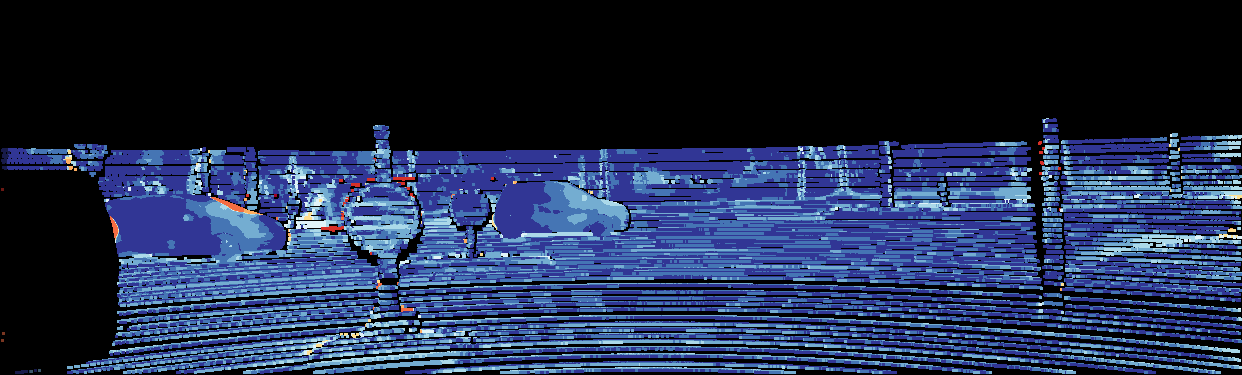}
&\includegraphics[width=0.24\linewidth]{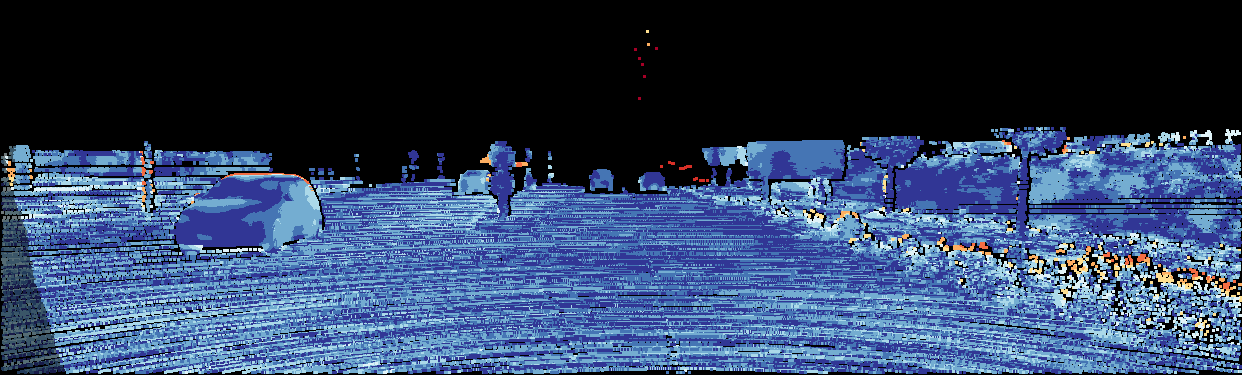}\\
\raisebox{2px}{\rotatebox{90}{\small PSM-Net}}
\includegraphics[width=0.24\linewidth]{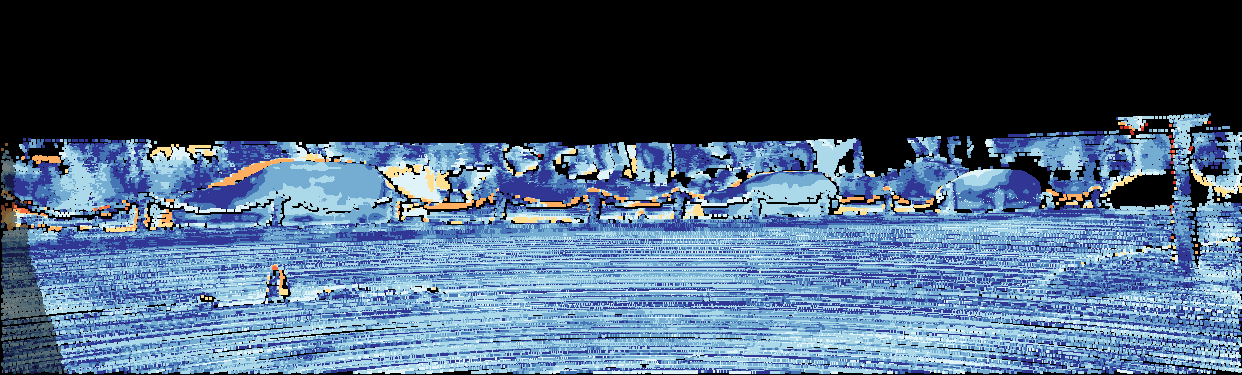}
&\includegraphics[width=0.24\linewidth]{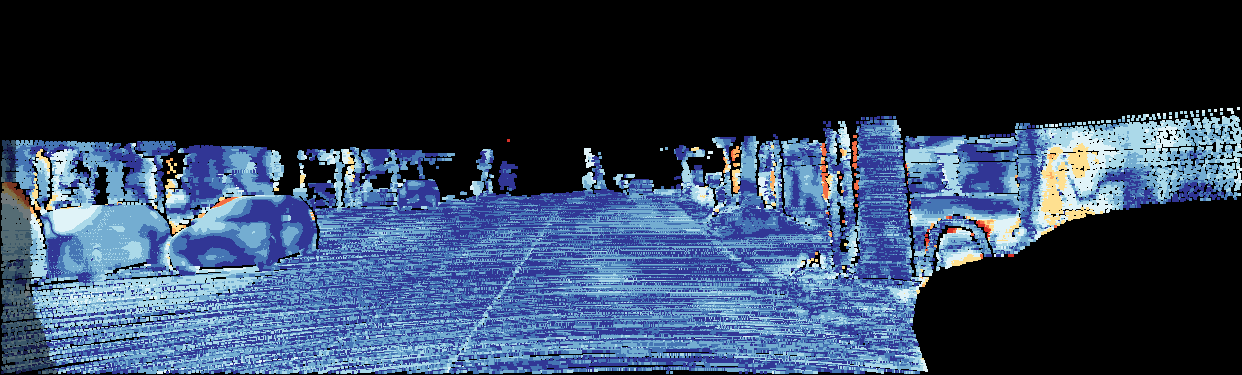}
&\includegraphics[width=0.24\linewidth]{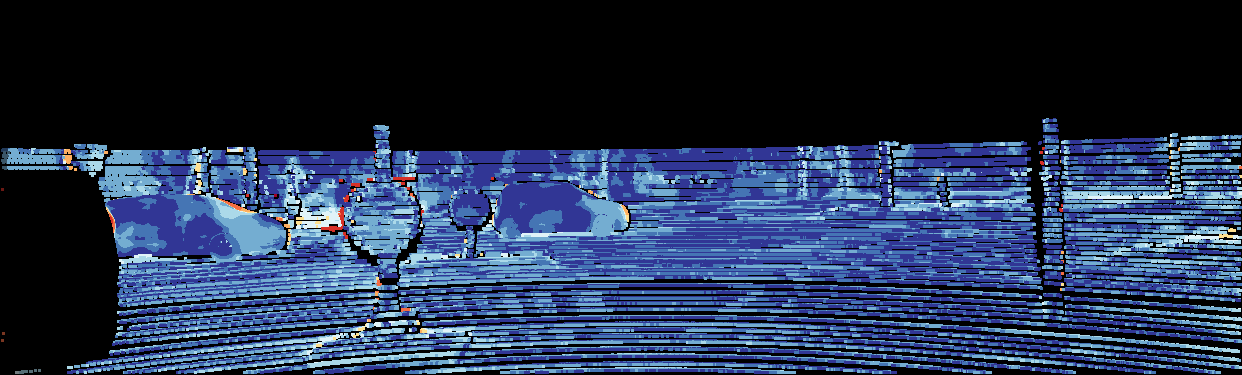}
&\includegraphics[width=0.24\linewidth]{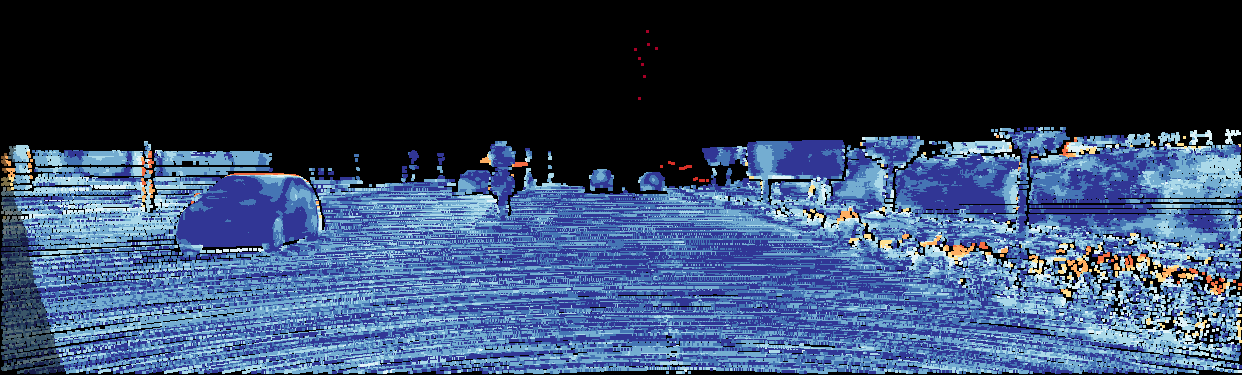}\\
\raisebox{2px}{\rotatebox{90}{\small DispNet}}
\includegraphics[width=0.24\linewidth]{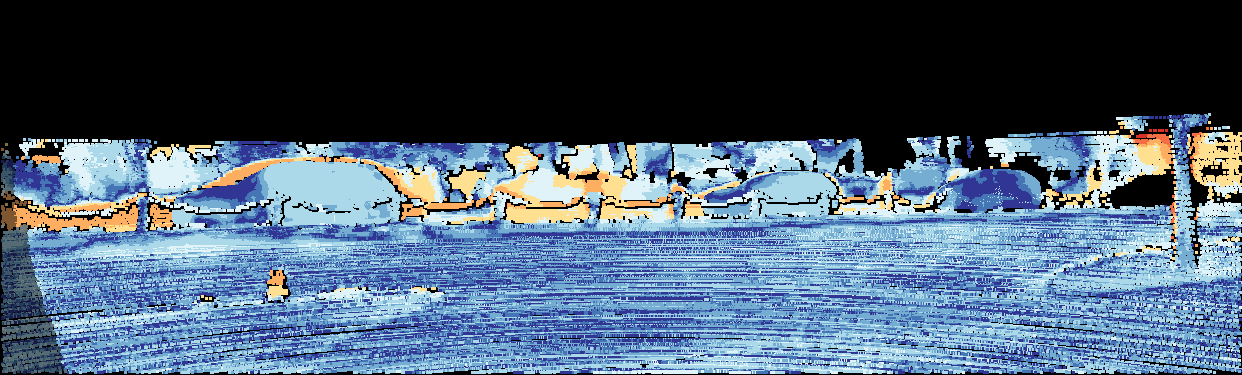}
&\includegraphics[width=0.24\linewidth]{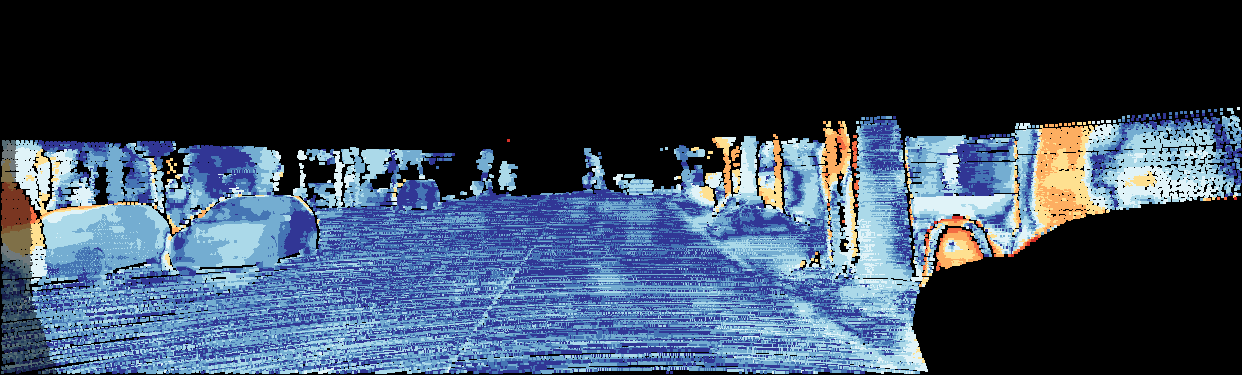}
&\includegraphics[width=0.24\linewidth]{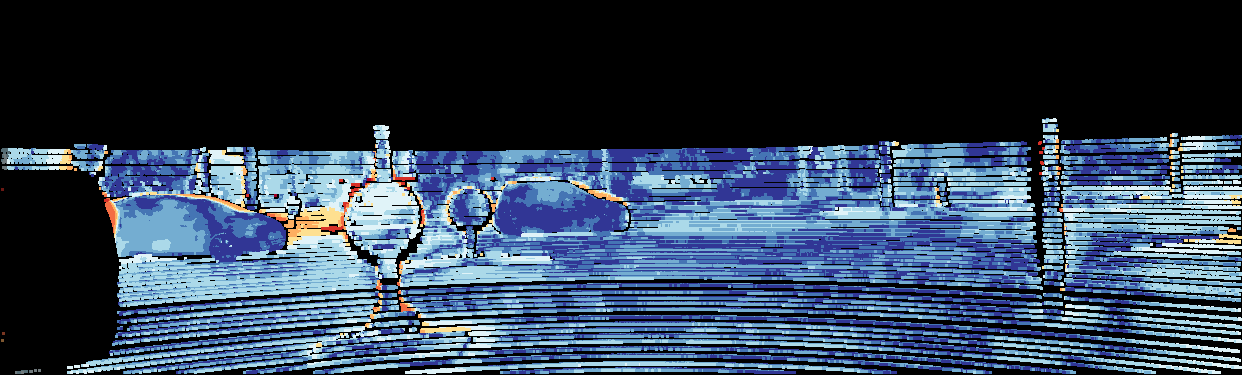}
&\includegraphics[width=0.24\linewidth]{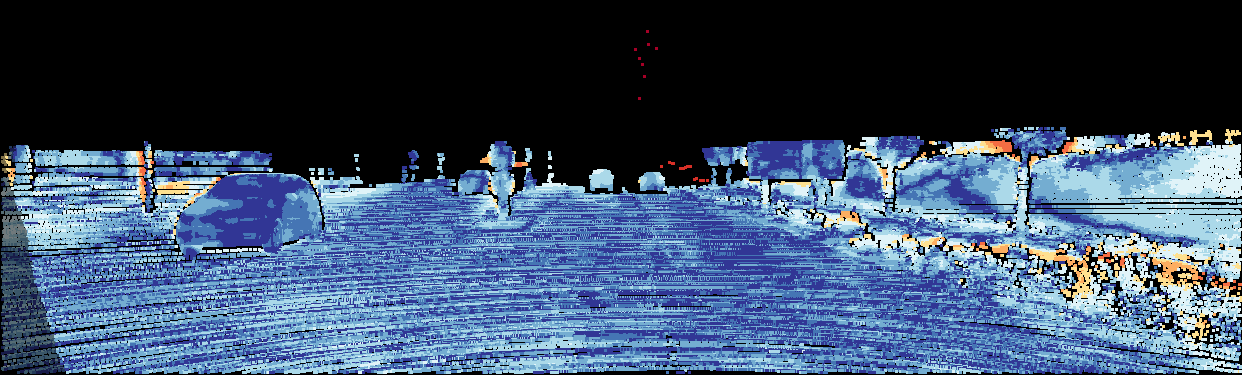}\\
\raisebox{1px}{\rotatebox{90}{\small Our-Fast}}
\includegraphics[width=0.24\linewidth]{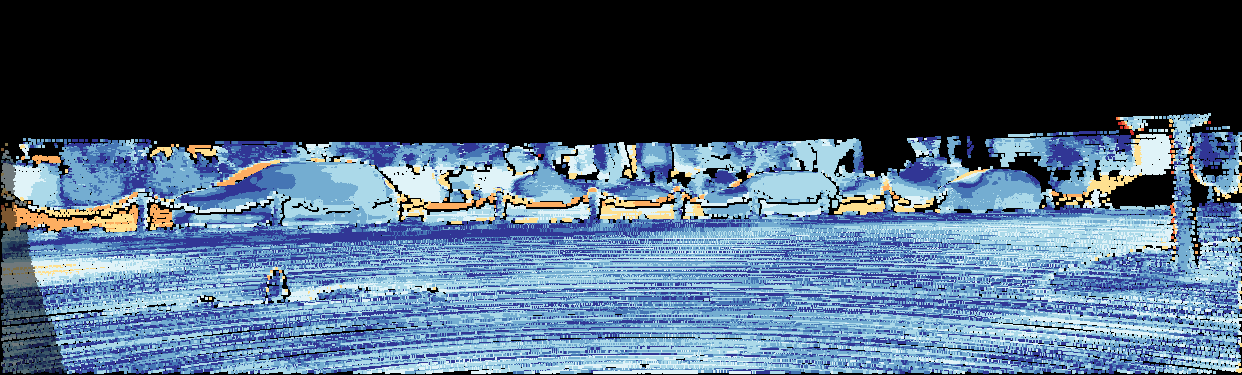}
&\includegraphics[width=0.24\linewidth]{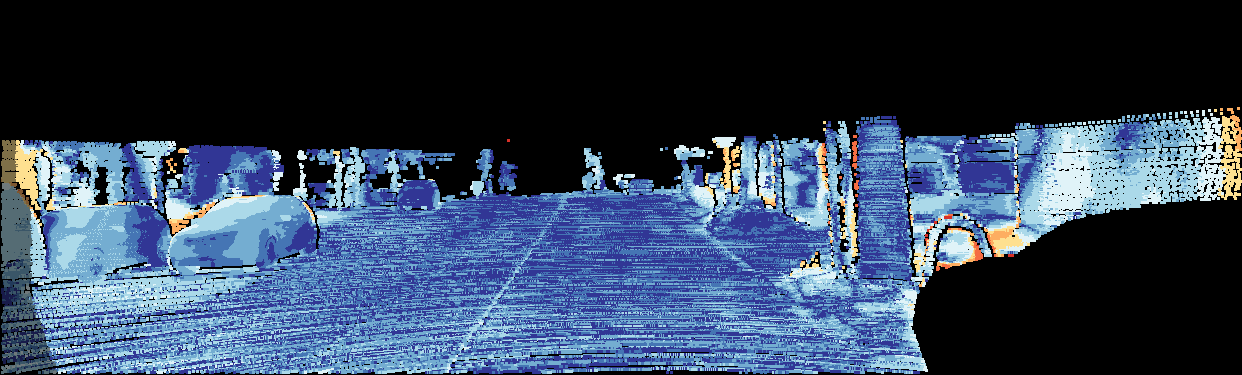}
&\includegraphics[width=0.24\linewidth]{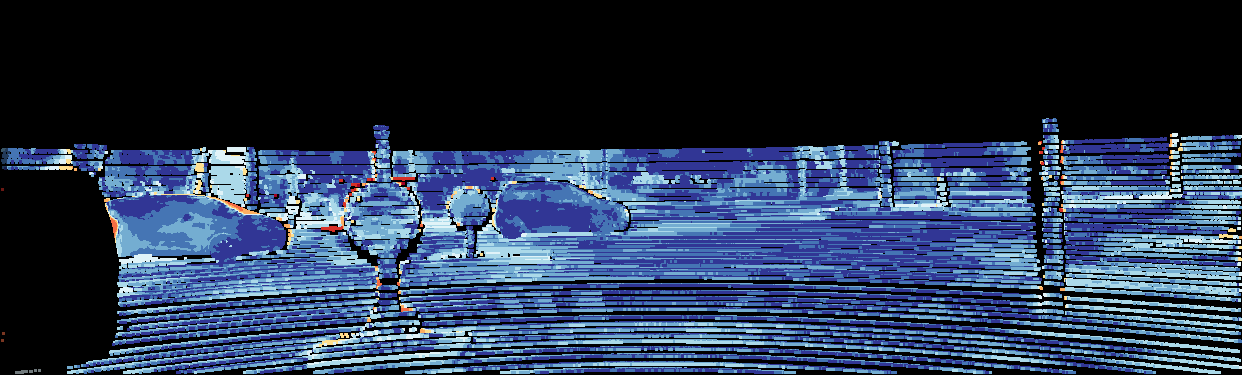}
&\includegraphics[width=0.24\linewidth]{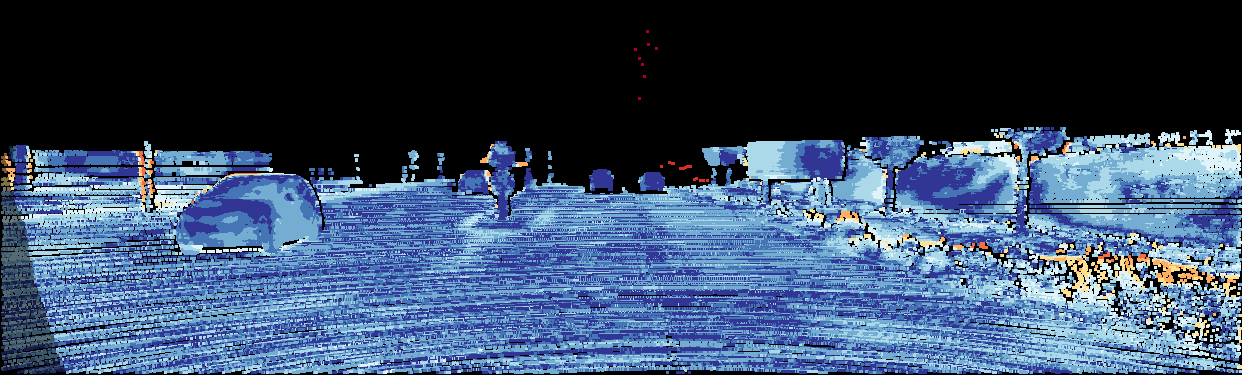}\\
\raisebox{1px}{\rotatebox{90}{\small Our-Best}}
\includegraphics[width=0.24\linewidth]{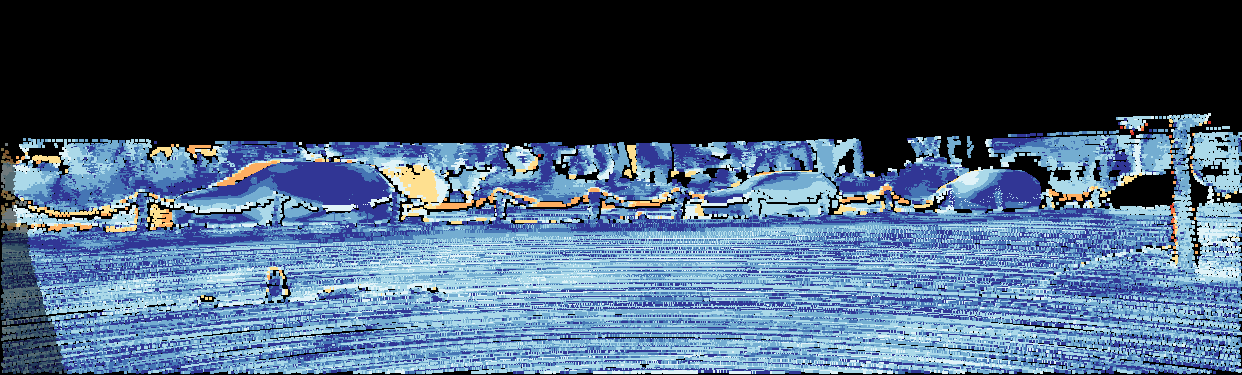}
&\includegraphics[width=0.24\linewidth]{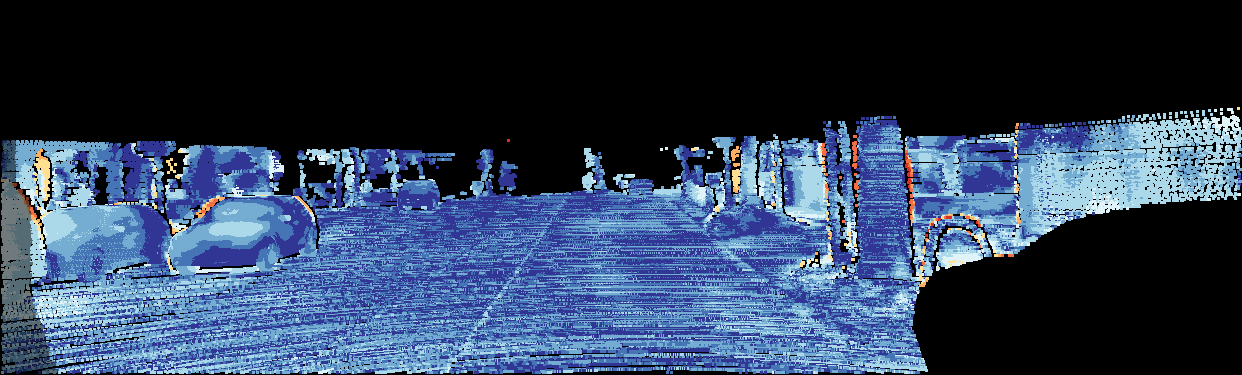}
&\includegraphics[width=0.24\linewidth]{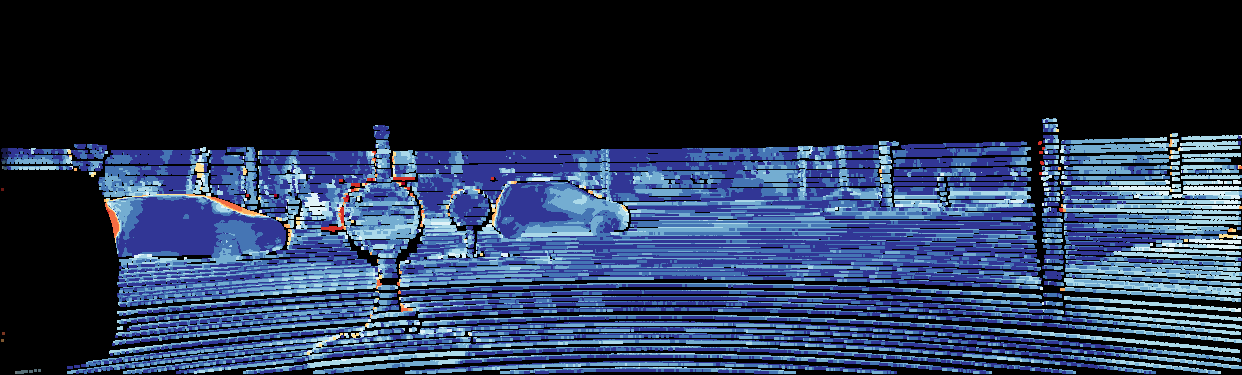}
&\includegraphics[width=0.24\linewidth]{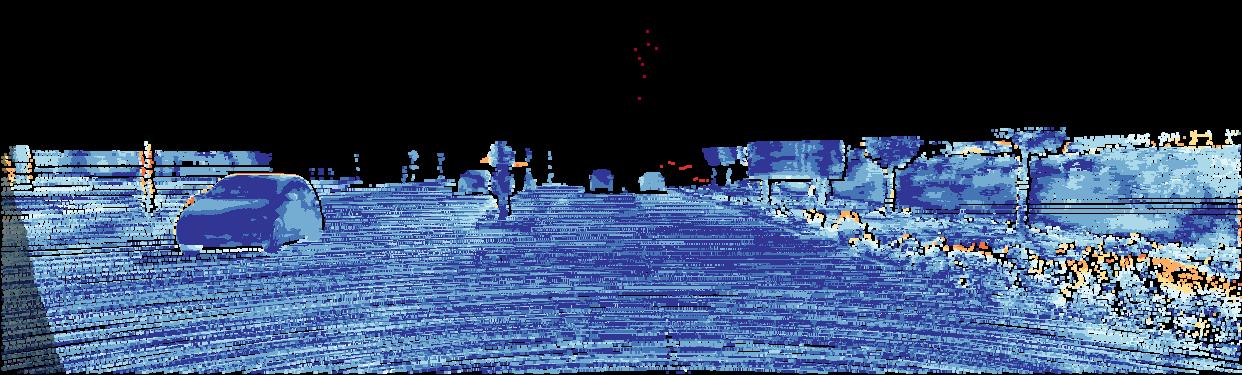}
\end{tabular}
\caption{\textbf{Qualitative Results on KITTI 2015 Test Set.}  
\weichiu{Orange corresponds to erroneous prediction.}}
\label{fig:qual-results}
\end{figure*}

\section{Experiments}
We compare our approach against the best performing algorithms \cite{zbontar2016stereo, kendall2017end,cheng2018learning, pang2017cascade, song2018edgestereo, chang2018pyramid} and the real-time models \cite{Tonioni_2019_CVPR, mayer2016large}. 
 
Specifically, we evaluate two variants of our method, namely \texttt{DeepPruner-Best} and \texttt{DeepPruner-Fast}. \texttt{DeepPruner-Best} downsamples the cost volume by 4 times while \texttt{DeepPruner-Fast} downsamples it by 8. The rest of the two models remain the same. 

In this section, we first describe our experimental setup. Next we evaluate our approach on challenging public benchmarks. Finally we comprehensively study the characteristic of our model.

\begin{table*}[tb]
\centering
\scalebox{.9}{
\begin{tabular}{cccccccccccccc}
\specialrule{.2em}{.1em}{.1em}

\multicolumn{6}{c}{Network Component}& Inference&\multicolumn{3}{c}{KITTI 2015 (\%)} & SceneFlow\\
Feat. Extr.& PM-1 &CRP &PM-2$^\ast$ &CA &RefineNet & Runtime  &\emph{bg}&\emph{fg}&\emph{all} & EPE \\
\hline
\checkmark& \checkmark& & &\checkmark &\checkmark & 120 ms & 2.05 & 3.57 & 2.28 & 0.982\\

\checkmark& \checkmark& \checkmark & &\checkmark & \checkmark & 172 ms & 1.65 & 3.27 & 1.90 & 0.868\\
\checkmark& \checkmark& \checkmark &\checkmark & \checkmark & &178 ms& 2.04 & 3.86 & 2.32 & 1.283\\

\checkmark& \checkmark& \checkmark &\checkmark& \checkmark & \checkmark & 182 ms & 1.61 & 2.90 & 1.8 & 0.858\\
\specialrule{.1em}{.05em}{.05em}
\end{tabular}}
\caption{\textbf{Contributions of each network component:} \weichiu{Confidence range predictor (CRP) and PatchMatch (PM) significantly prune out the solution space, thus enable better cost aggregation (CA). RefineNet further improves the estimation by incoporating more visual guidance.
({$^\ast$}: PM-2 is only used during SceneFlow pre-training.) }}
\label{tab:ablation-architecture}
\end{table*}

\begin{figure*}[tb]
\centering
\def\arraystretch{0.6}
\setlength{\tabcolsep}{1pt}
\begin{tabular}{ccccc}
RGB & Disparity & Error & Uncertainty \\
\raisebox{19px}{\rotatebox{90}{}}
\includegraphics[width=0.24\linewidth]{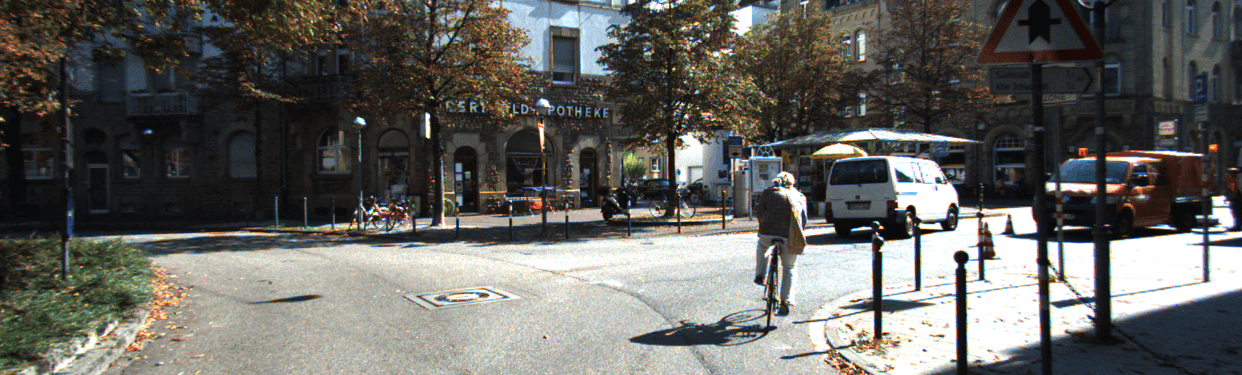}
&\includegraphics[width=0.24\linewidth]{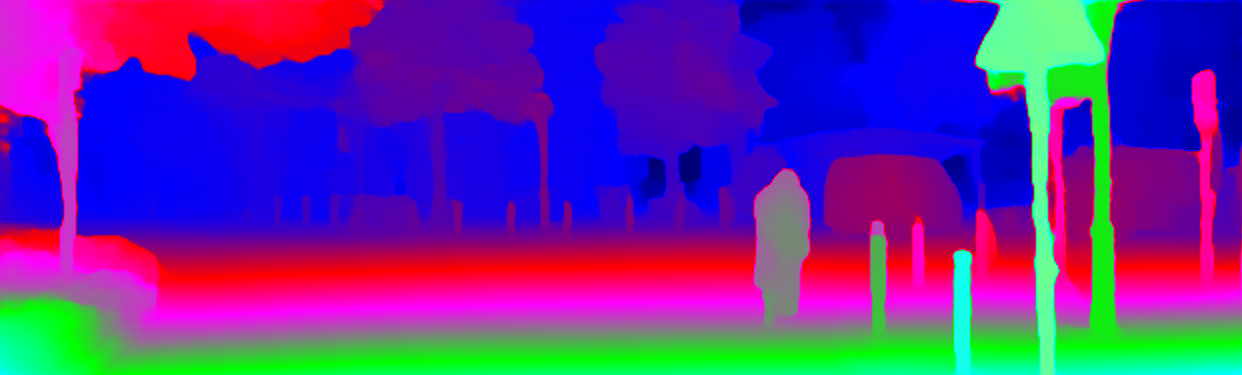}
&\includegraphics[width=0.24\linewidth]{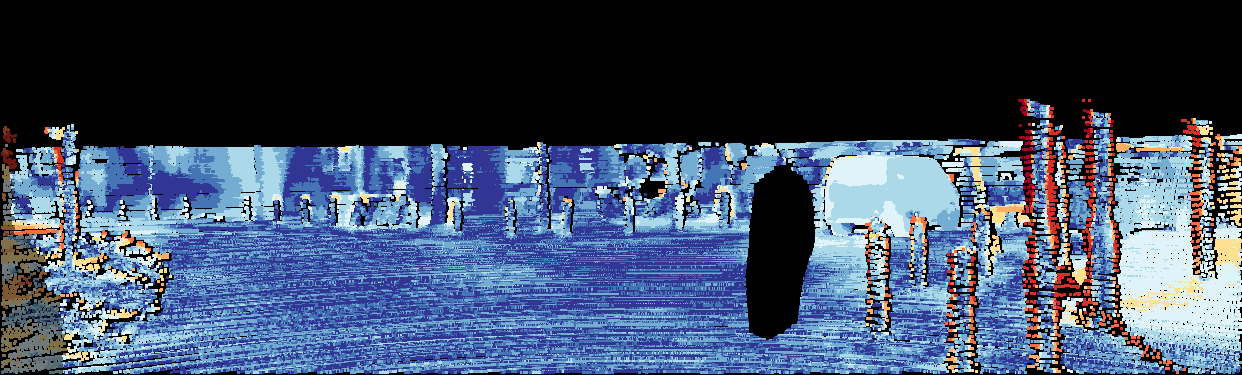}
&\includegraphics[width=0.24\linewidth]{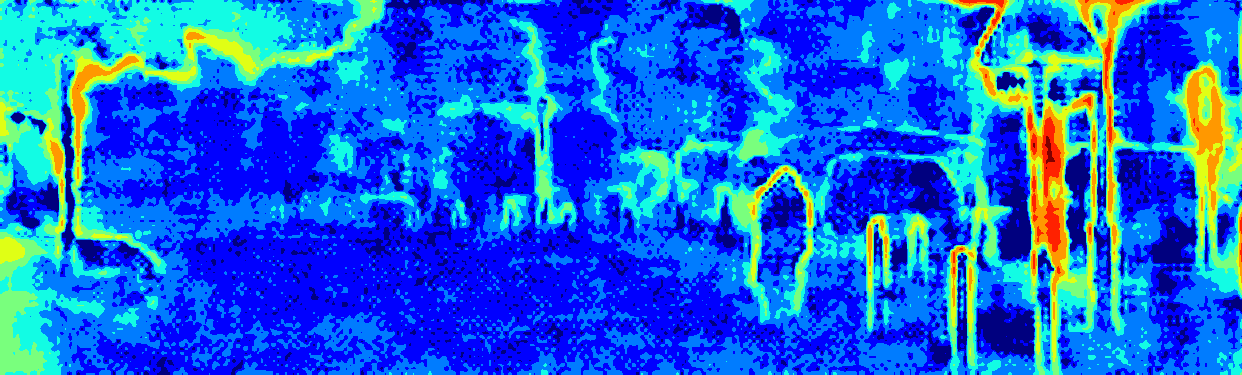}\\
\raisebox{17px}{\rotatebox{90}{}}
\includegraphics[width=0.24\linewidth]{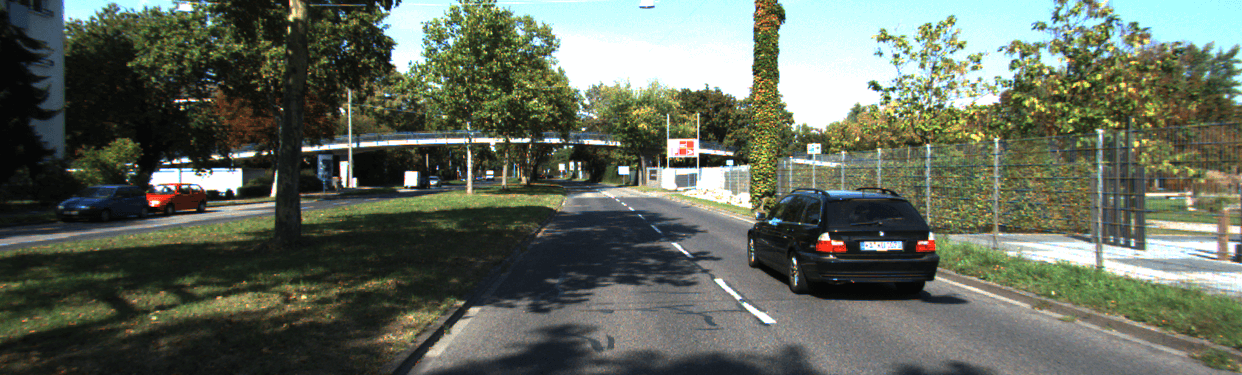}
&\includegraphics[width=0.24\linewidth]{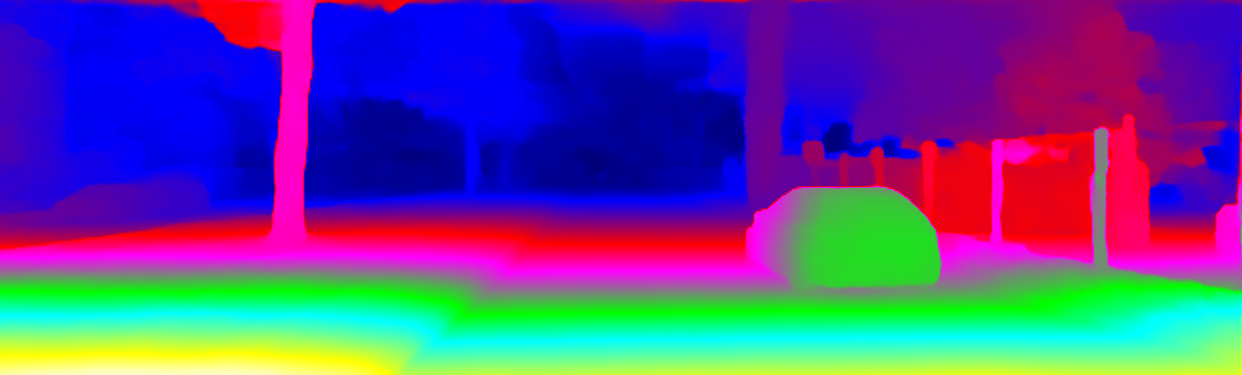}
&\includegraphics[width=0.24\linewidth]{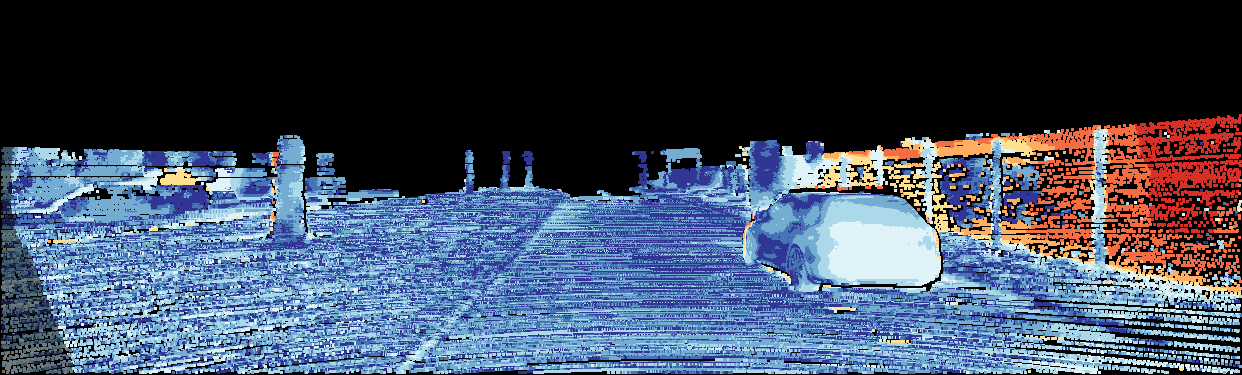}
&\includegraphics[width=0.24\linewidth]{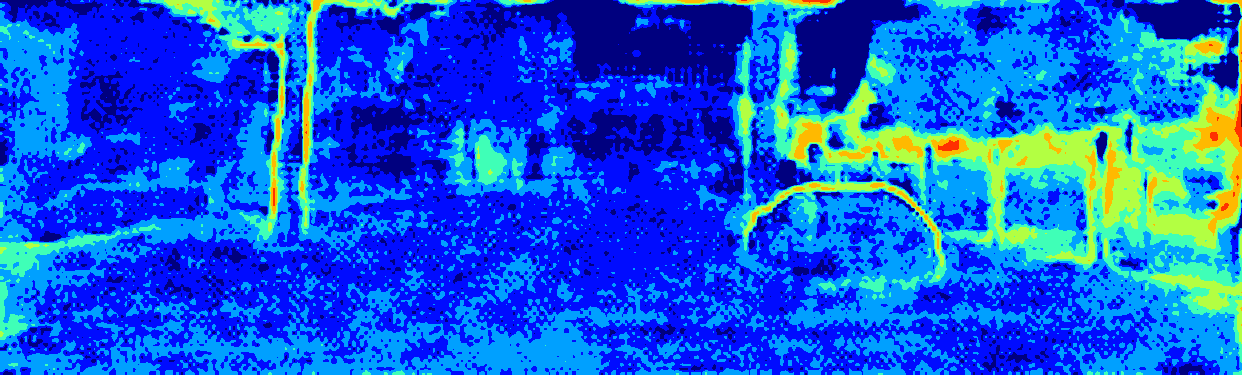}\\
\raisebox{12px}{\rotatebox{90}{}}
\includegraphics[width=0.24\linewidth]{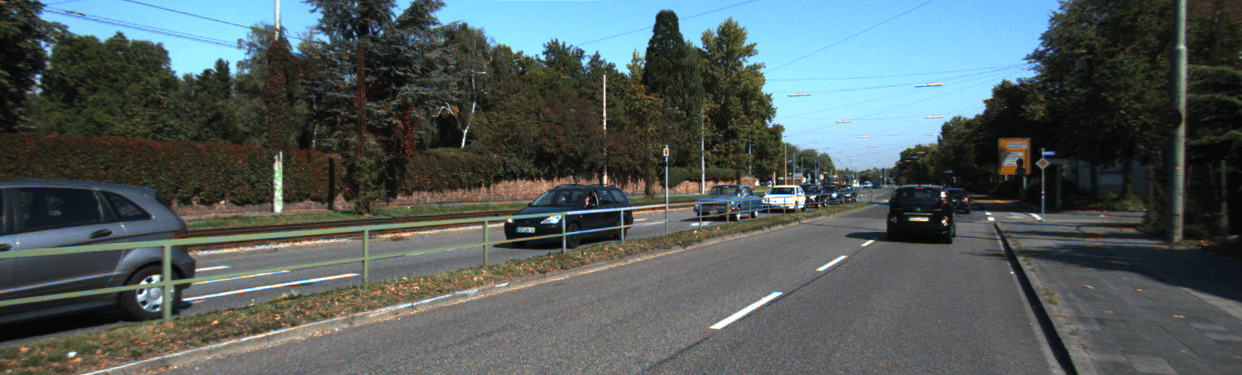}
&\includegraphics[width=0.24\linewidth]{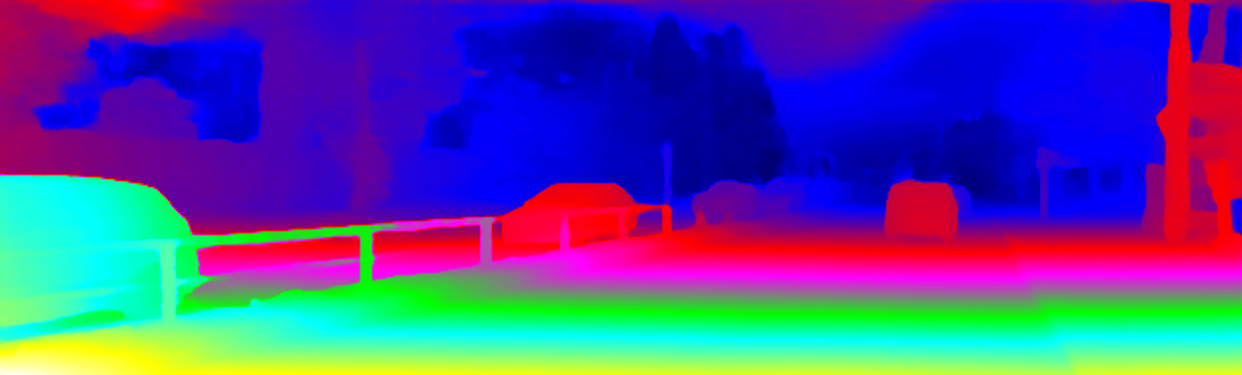}
&\includegraphics[width=0.24\linewidth]{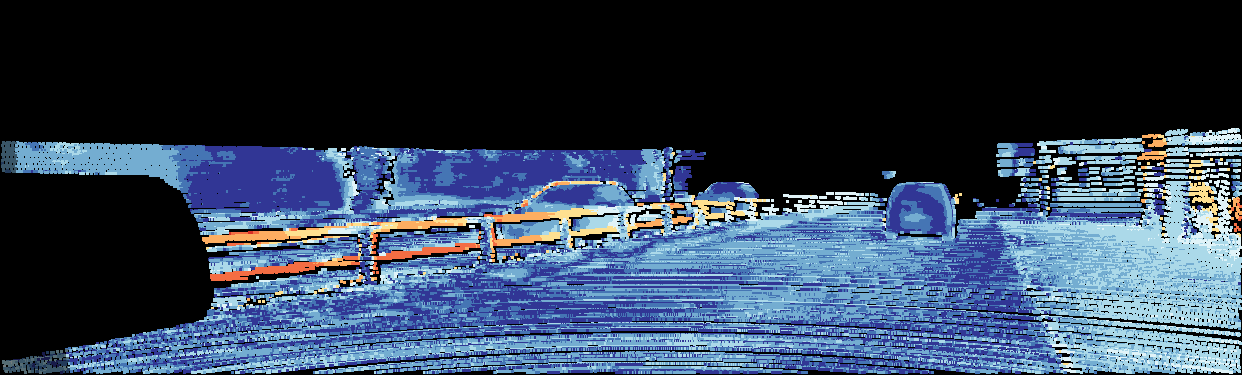}
&\includegraphics[width=0.24\linewidth]{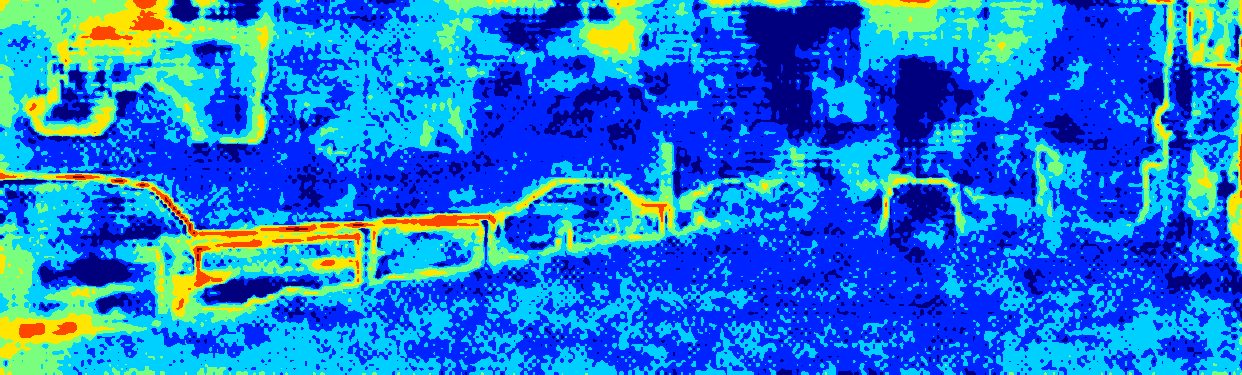}\\
\raisebox{17px}{\rotatebox{90}{}}
\includegraphics[width=0.24\linewidth]{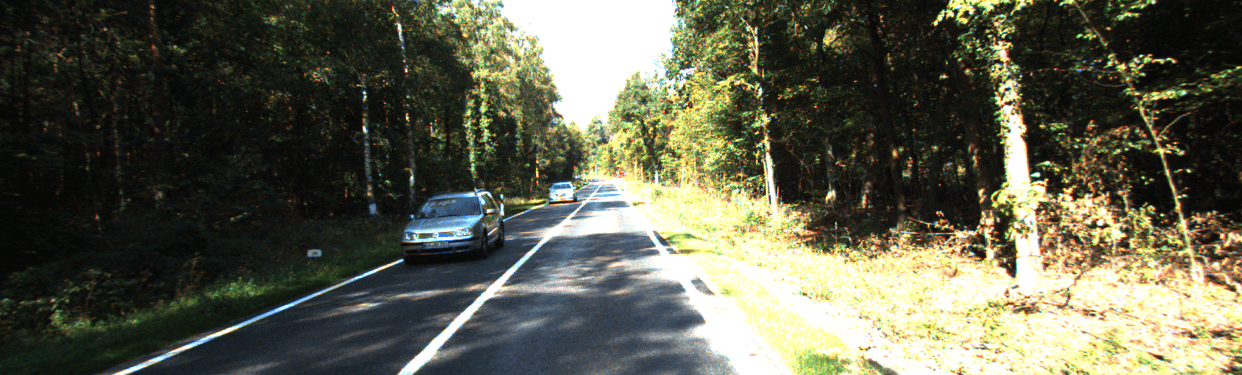}
&\includegraphics[width=0.24\linewidth]{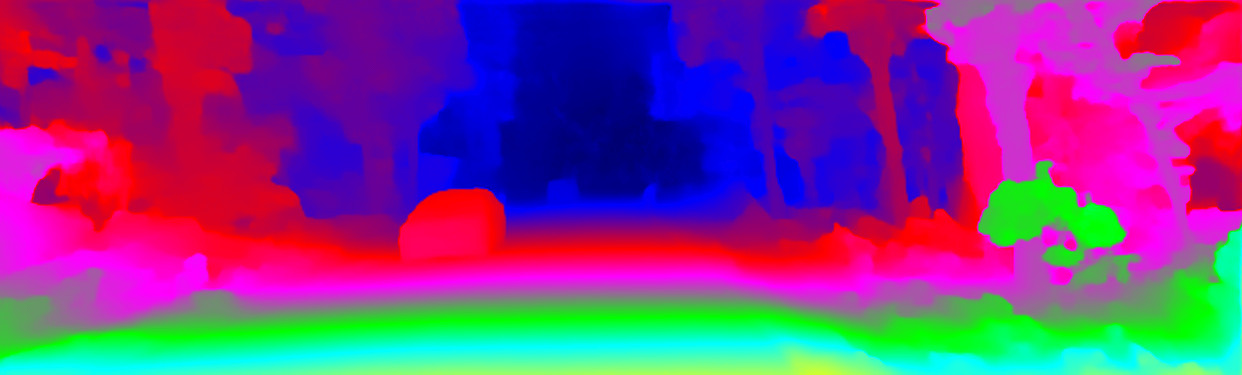}
&\includegraphics[width=0.24\linewidth]{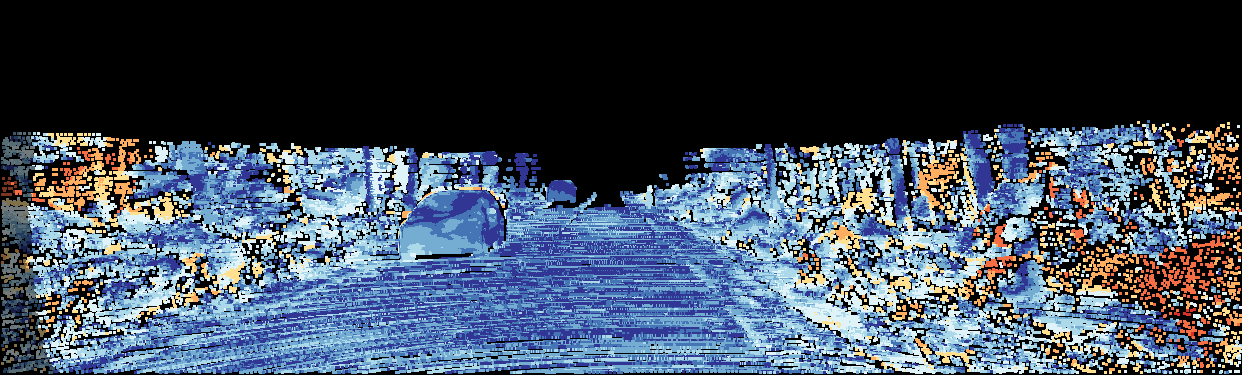}
&\includegraphics[width=0.24\linewidth]{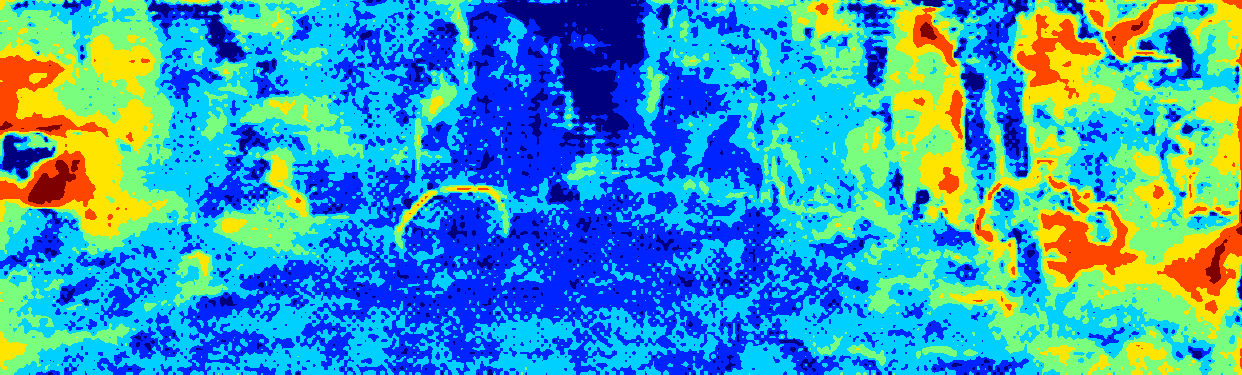}\\
\\\end{tabular}
\caption{\textbf{Uncertainty vs Error.} As can be seen, uncertainty correlates highly with the error, thereby being a good indicator of the potential errors. }
\label{fig:uncertainty}
\end{figure*}

\subsection{Dataset}

\paragraph{SceneFlow:} As proposed in \cite{mayer2016large}, this is a synthetic dataset consisting of dense ground truth disparity maps for 35454 training and 4370 testing stereo pairs with dimensions as (H=540,W=960).  
We used End-Point-Error (EPE) as the evaluation metric for SceneFlow dataset. 
\vspace{-3mm}

\paragraph{KITTI 2015:} This is a real world dataset consisting of 200 training and 200 testing stereo pairs with dimensions (H=376,W=1240). The ground truth disparities are obtained from Lidar points collected using Velodyne HDL-64E laserscanner. Unlike the prior version KITTI 2012, dense ground disparities are present for the dynamic scenes. The evaluation metrics used are same as that provided by the benchmark i.e percentage of outliers.

\subsection{Implementation Details}

The model was trained end-to-end on randomly cropped image patches of size (H=256,W=512) using Adam($\beta_{1}$=0.9, $\beta_{2}$=0.999) as the optimizer. Color normalization was done on the input images using ImageNet statistics(mean and std) as the sole pre-processing step. All models were trained on 4 Nvidia-TitanXp GPUs.  For all datasets, we use the hyper-parameter $\lambda = 0.315$, $\gamma = 2.4$ in our loss function. We only computed loss over pixels with ground truth disparities between 0 and 192 during training. Evaluation is conducted over the all pixels regardless of their disparity values.

For sceneflow dataset, we trained the model from scratch with a batch size of 16 for 64 epochs. The initial learning rate was set to 0.001 and was decayed by 0.0003 after every 20 epochs. 

For KITTI dataset, we combined KITTI 2012 and KITTI 2015 image pairs, resulting in a total of 394 training image pairs. \shivam{We reserved out 40 images from the total 394 images for validation.}

We then used the pre-trained Sceneflow model and finetuned it for another 1040 epochs. After 800 epochs, all batch norm layers were switched to eval mode, i.e the running mean and std statistics were kept fixed for further training. We used a batch size of 16 for \texttt{DeepPruner-Best} model, while \texttt{DeepPruner-Fast} model was trained with a batch size of 64. We used an initial lr of 0.0001 and reduced it once, after 500 epochs to 0.00005.
For submission to the KITTI test benchmark, we re-trained the model on all the 394 training images for 1040 epochs.

\subsection{Experimental Results}

\paragraph{SceneFlow:} \weichiu{As shown in Tab.~\ref{tab:quant-scenflow}, our method outperforms most approaches by a large margin and achieves the second best results. Comparing to state of the art \cite{cheng2018learning}, our best model is 2.5 times faster and our fast model is over 8 times faster. Comparing against the real-time approach \cite{mayer2016large}, our method reduces the end-point-error by $40\%$ with almost the same runtime. Fig.~\ref{fig:qual-scene} depicts the qualitative results. DeepPruner captures both large disparities and small objects, and is able to produce sharp estimation on image boundaries.}

\paragraph{KITTI:} \weichiu{Tab.~\ref{tab:quant-2015} showcases the performance and runtime of all competing algorithms on KITTI stereo benchmark.  \texttt{DeepPruner-Best} achieves comparable performance to state-of-the-art approaches while being significantly faster. 
Comparing to real-time stereo models, \texttt{DeepPruner-Fast} reduces the outlier ratio by more than $40\%$. Fig.~\ref{fig:qual-results} visualize a few stereo results on test set. DeepPruner produces competitive estimation among various scenarios.}

\subsection{Analysis}

\paragraph{Ablation study:}
\weichiu{To understand the effectiveness of each component in DeepPruner, we evaluate our model with different configurations. 
As shown in Tab.~\ref{tab:ablation-architecture}, confidence range predictor is of crucial importance to our model. With the help of refinement network, we can further capture the sharp edges as well as the fine-grained details, and improve the overall stereo estimation.}

\paragraph{Visualizing confidence range predictor:} \weichiu{
The goal of confidence range predictor is to prune out the space of unlikely matches and ensure the expensive cost volume operations only happen at a few disparity values. 
To understand the efficacy of the predictor, we visualize the predicted search range as well as the GT disparity for pixels along a horizontal line. As shown in Fig.~\ref{fig:qual-uncertainty}, our confident range is quite small in most cases, which greatly reduce the computation and memory burden of cost volume prediction.}

\paragraph{Uncertainty:}
\weichiu{The range prediction can also be considered as a measurement of confidence/uncertainty level --- the larger the range, the more uncertain the model is. To validate this hypothesis, we compare the predicted confidence range (\emph{i.e.,} max range minus min range) against the disparity error map over several validation images. As shown in Fig.~\ref{fig:uncertainty}, the uncertainty (predicted range) map and the error map are highly correlated, suggesting that it could be a good indicator of the potential errors. 
To further verify this, we track the change in the metric by gradually removing uncertain pixels, starting from most uncertain ones.
By removing 6\% of uncertain pixels, we improve the outlier ratio by $38\%$. 
This clearly indicates that our high confidence regions have very low error whereas most of the error happens at low confidence regions. }

\begin{table}[h]
\centering
\scalebox{0.6}{
\begin{tabular}{lcccccccccccc}
\specialrule{.2em}{.1em}{.1em}
Method & \multicolumn{2}{c}{KITTI2015} & \multicolumn{2}{c}{MiddleburyV3} & \multicolumn{2}{c}{ETH3D} & ROB\\
& Rank & D1-all (3px) & Rank & 4x & Rank & 4x & Overall-Rank\\ 
\hline
PSMNet\_ROB \cite{chang2018pyramid} & 2 & 2.31 & 11 & 29.2 & 6 & 0.54 & 5 \\
DN-CSS\_ROB \cite{ROB} & 8 & 2.94 & 3 & \textbf{19.6} & 2 & 0.38 & 3\\
iResNet\_ROB \cite{iResNet} & 3 & 2.71 & 6 & 22.1 & 3 & 0.40 & 2\\
DeepPruner\_ROB & \textbf{1} & \textbf{2.23} & 7 & 21.9 & \textbf{1} & \textbf{0.34} & \textbf{1}\\

\specialrule{.1em}{.05em}{.05em} \\
\end{tabular}
}
\caption{\weichiu{\textbf{Comparison against top 3 approaches on Robust Vision Challenge.} \slwang{Our proposed approach achieves the highest overall rank.}}}
\label{tab:rob}
\vspace{-6mm}
\end{table}

\vspace{-2mm}

\begin{figure}[h]
\centering
\def\arraystretch{0.6}
\setlength{\tabcolsep}{1pt}
\begin{tabular}{ccccc}
{\small Image} &{\small iResNet \cite{iResNet}} &{\small DN-CSS \cite{ROB}} &{\small Our-Best}\\
\raisebox{2px}{\rotatebox{90}{\footnotesize{Middlebury}}}
\includegraphics[width=0.22\linewidth, height=0.17\linewidth]{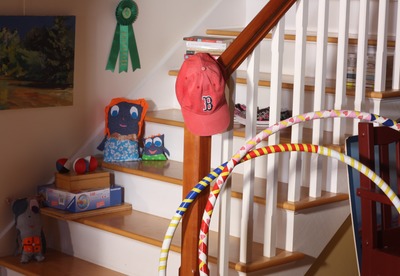}
&\includegraphics[width=0.22\linewidth, height=0.17\linewidth]{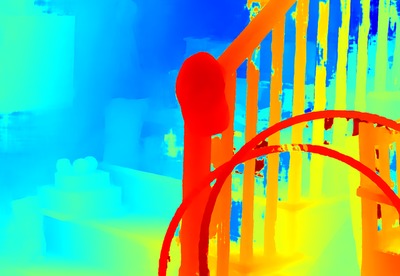}
&\includegraphics[width=0.22\linewidth, height=0.17\linewidth]{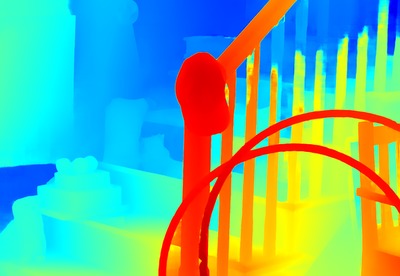}
&\includegraphics[width=0.22\linewidth, height=0.17\linewidth]{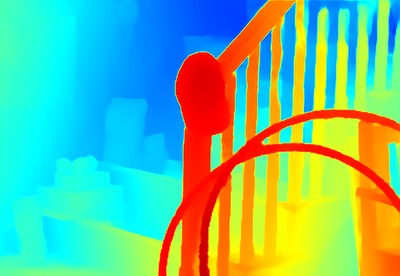}\\
\raisebox{6px}{\rotatebox{90}{\footnotesize{ETH3D}}}
\includegraphics[width=0.22\linewidth, height=0.17\linewidth]{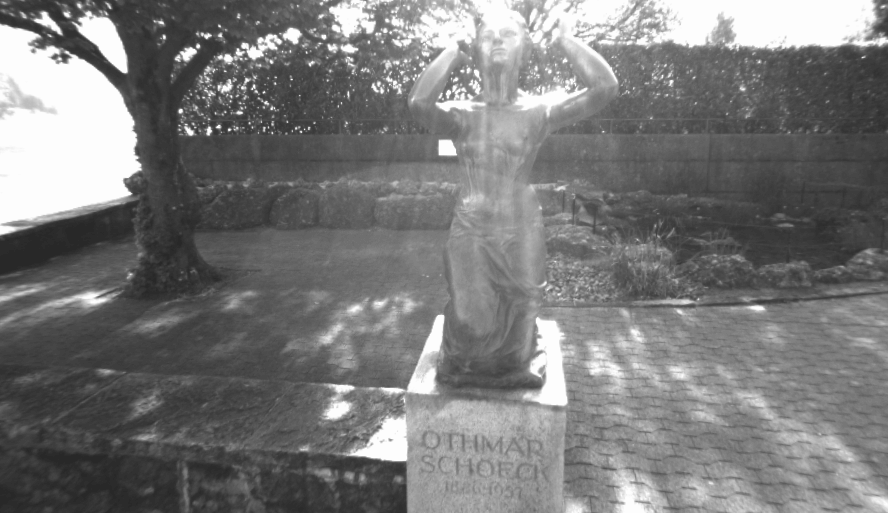}
&\includegraphics[width=0.22\linewidth, height=0.17\linewidth]{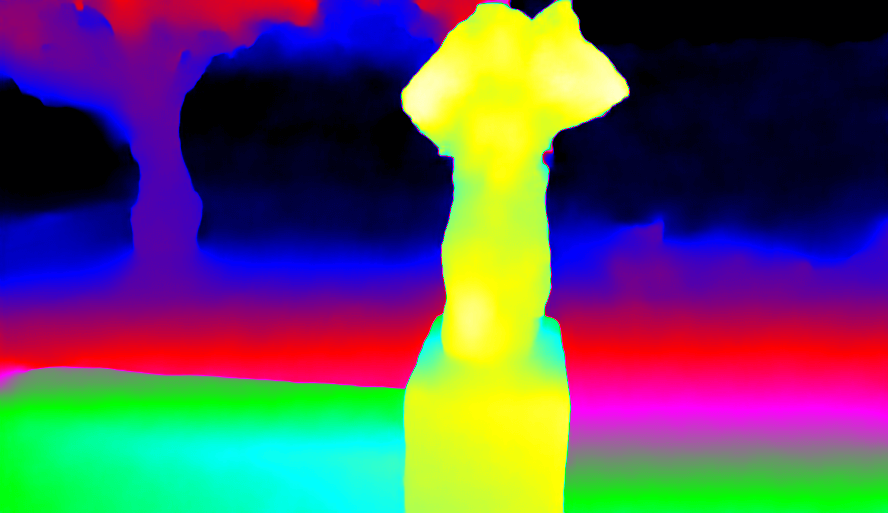}
&\includegraphics[width=0.22\linewidth, height=0.17\linewidth]{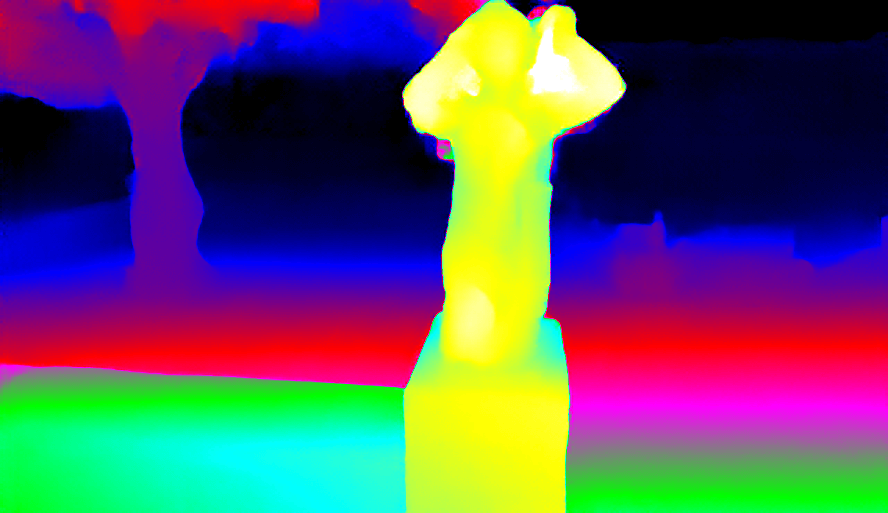}
&\includegraphics[width=0.22\linewidth, height=0.17\linewidth]{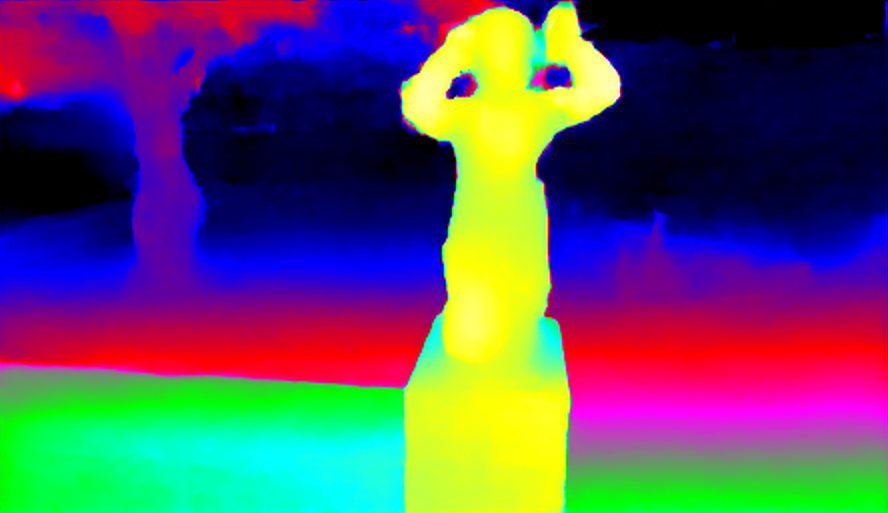}

\end{tabular}
\caption{\textbf{Qualitative Results on ROB Challenge:} 
\weichiu{Our method captures the fine-grained geometry of the scenes. See the stair armrest and the arms of the sculpture.}}
\label{fig:rob}
\end{figure}

 \paragraph{Robustness and generalizability:}
 \weichiu{To corroborate how well our model generalizes across different scenarios, we evalute our model on the Robust Vision Challenge \cite{ROB}. Specifically, we fine-tune our sceneflow-pretrained model on KITTI \cite{menze2015object}, ETH3D\cite{schoeps2017cvpr}, and MiddleburyV3 \cite{conf/dagm/ScharsteinHKKNWW14} jointly and report results on all three datasets. As shown in Tab. \ref{tab:rob}, DeepPruner achieves the highest rank on two datasets and ranks first on the overall ranking. It is also able to capture the fine-grained geometry of various scenes (see Fig. \ref{fig:rob}). }

\vspace{-4mm}

\paragraph{Runtime and memory analysis:} \weichiu{We benchmark the runtime of each component in the model during inference in Tab. \ref{tab:runtime}. As PatchMatch and confidence range predictor progressively reduce the possible solution space, we only need to perform cost aggregation among a small subset of disparities. The model is thus significantly faster.}
 \weichiu{To further demonstrate the efficiency of our model, we compare our memory consumption against previous full cost volume approach \cite{chang2018pyramid}. For a pair of full-size KITTI stereo images, PSM-Net \cite{chang2018pyramid} takes up to {4351} MB memory during inference. In contrast, our \texttt{DeepPruner-Best} and \texttt{DeepPruner-Fast} merely consume {1161} MB and 805 MB memory respectively. The storage requirements are less than one fourth of \cite{chang2018pyramid}, which shows the potential to be integrated in mobile computing platform.}

\section{Conclusion}

In this paper we show how to exploit the fact that we can quickly prune parts of the cost volume for each pixel without requiring to fully evaluate its matching score. 
Towards this goal, we developed an end-to-end trainable network that exploits a novel differentiable PatchMatch as part of its internal structure. 
Our experiments show that our model achieves best performance among real-time methods and comparable results with the best performing methods while remaining several times faster.
In the future we plan to apply our approach to both optical flow and scene flow tasks.

{\small
\bibliographystyle{ieee_fullname}
\bibliography{egbib}

\begin{thebibliography}{10}\itemsep=-1pt

\bibitem{ROB}
http://www.robustvision.net/.

\bibitem{barnard1982computational}
Stephen~T Barnard and Martin~A Fischler.
\newblock Computational stereo.
\newblock Technical report, SRI INTERNATIONAL MENLO PARK CA ARTIFICIAL
  INTELLIGENCE CENTER, 1982.

\bibitem{barnes2009patchmatch}
Connelly Barnes, Eli Shechtman, Adam Finkelstein, and Dan~B Goldman.
\newblock Patchmatch: A randomized correspondence algorithm for structural
  image editing.
\newblock In {\em TOG}. ACM, 2009.

\bibitem{generalizedpm}
Connelly Barnes, Eli Shechtman, Dan~B Goldman, and Adam Finkelstein.
\newblock The generalized {PatchMatch} correspondence algorithm.
\newblock In {\em ECCV}, Sept. 2010.

\bibitem{besse2014pmbp}
Frederic Besse, Carsten Rother, Andrew Fitzgibbon, and Jan Kautz.
\newblock Pmbp: Patchmatch belief propagation for correspondence field
  estimation.
\newblock {\em IJCV}, 2014.

\bibitem{bleyer2011patchmatch}
Michael Bleyer, Christoph Rhemann, and Carsten Rother.
\newblock Patchmatch stereo-stereo matching with slanted support windows.

\bibitem{calonder2010brief}
Michael Calonder, Vincent Lepetit, Christoph Strecha, and Pascal Fua.
\newblock Brief: Binary robust independent elementary features.
\newblock In {\em ECCV}, 2010.

\bibitem{chang2018pyramid}
Jia-Ren Chang and Yong-Sheng Chen.
\newblock Pyramid stereo matching network.
\newblock In {\em CVPR}, 2018.

\bibitem{cheng2018learning}
Xinjing Cheng, Peng Wang, and Ruigang Yang.
\newblock Learning depth with convolutional spatial propagation network.
\newblock {\em arXiv}, 2018.

\bibitem{ferstl2013image}
David Ferstl, Christian Reinbacher, Rene Ranftl, Matthias R{\"u}ther, and Horst
  Bischof.
\newblock Image guided depth upsampling using anisotropic total generalized
  variation.
\newblock In {\em ICCV}, pages 993--1000, 2013.

\bibitem{Geiger2012CVPR}
Andreas Geiger, Philip Lenz, and Raquel Urtasun.
\newblock Are we ready for autonomous driving? the kitti vision benchmark
  suite.
\newblock In {\em CVPR}, 2012.

\bibitem{he2012computing}
Kaiming He and Jian Sun.
\newblock Computing nearest-neighbor fields via propagation-assisted kd-trees.
\newblock In {\em CVPR}, 2012.

\bibitem{he2014spatial}
Kaiming He, Xiangyu Zhang, Shaoqing Ren, and Jian Sun.
\newblock Spatial pyramid pooling in deep convolutional networks for visual
  recognition.
\newblock In {\em ECCV}, 2014.

\bibitem{hirschmuller2008stereo}
Heiko Hirschmuller.
\newblock Stereo processing by semiglobal matching and mutual information.
\newblock {\em TPAMI}, 2008.

\bibitem{kendall2017end}
Alex Kendall, Hayk Martirosyan, Saumitro Dasgupta, Peter Henry, Ryan Kennedy,
  Abraham Bachrach, and Adam Bry.
\newblock End-to-end learning of geometry and context for deep stereo
  regression.
\newblock 2017.

\bibitem{khamis2018stereonet}
Sameh Khamis, Sean Fanello, Christoph Rhemann, Adarsh Kowdle, Julien Valentin,
  and Shahram Izadi.
\newblock Stereonet: Guided hierarchical refinement for real-time edge-aware
  depth prediction.
\newblock In {\em ECCV}, 2018.

\bibitem{korman2016coherency}
Simon Korman and Shai Avidan.
\newblock Coherency sensitive hashing.
\newblock {\em TPAMI}, 2016.

\bibitem{iResNet}
Zhengfa Liang, Yiliu Feng, Yulan Guo, Hengzhu Liu, Wei Chen, Linbo Qiao, Li
  Zhou, and Jianfeng Zhang.
\newblock Learning for disparity estimation through feature constancy.
\newblock 2018.

\bibitem{lowe2004distinctive}
David~G Lowe.
\newblock Distinctive image features from scale-invariant keypoints.
\newblock {\em IJCV}, 2004.

\bibitem{sift}
David~G Lowe et~al.
\newblock Object recognition from local scale-invariant features.
\newblock In {\em ICCV}, volume~99, pages 1150--1157, 1999.

\bibitem{lu2013patch}
Jiangbo Lu, Hongsheng Yang, Dongbo Min, and Minh~N Do.
\newblock Patch match filter: Efficient edge-aware filtering meets randomized
  search for fast correspondence field estimation.
\newblock In {\em CVPR}, 2013.

\bibitem{luo2016efficient}
Wenjie Luo, Alexander~G Schwing, and Raquel Urtasun.
\newblock Efficient deep learning for stereo matching.
\newblock In {\em CVPR}, 2016.

\bibitem{mayer2016large}
Nikolaus Mayer, Eddy Ilg, Philip Hausser, Philipp Fischer, Daniel Cremers,
  Alexey Dosovitskiy, and Thomas Brox.
\newblock A large dataset to train convolutional networks for disparity,
  optical flow, and scene flow estimation.
\newblock In {\em CVPR}, 2016.

\bibitem{menze2015object}
Moritz Menze and Andreas Geiger.
\newblock Object scene flow for autonomous vehicles.
\newblock In {\em CVPR}, 2015.

\bibitem{pang2017cascade}
Jiahao Pang, Wenxiu Sun, Jimmy~SJ Ren, Chengxi Yang, and Qiong Yan.
\newblock Cascade residual learning: A two-stage convolutional neural network
  for stereo matching.
\newblock In {\em ICCV Workshop on Geometry Meets Deep Learning}, Oct 2017.

\bibitem{conf/dagm/ScharsteinHKKNWW14}
Daniel Scharstein, Heiko Hirschmüller, York Kitajima, Greg Krathwohl, Nera
  Nesic, Xi Wang, and Porter Westling.
\newblock High-resolution stereo datasets with subpixel-accurate ground truth.
\newblock In {\em GCPR}, 2014.

\bibitem{scharstein2002taxonomy}
Daniel Scharstein and Richard Szeliski.
\newblock A taxonomy and evaluation of dense two-frame stereo correspondence
  algorithms.
\newblock {\em IJCV}, 2002.

\bibitem{schoeps2017cvpr}
Thomas Sch\"ops, Johannes~L. Sch\"onberger, Silvano Galliani, Torsten Sattler,
  Konrad Schindler, Marc Pollefeys, and Andreas Geiger.
\newblock A multi-view stereo benchmark with high-resolution images and
  multi-camera videos.
\newblock In {\em CVPR}, 2017.

\bibitem{song2018edgestereo}
Xiao Song, Xu Zhao, Hanwen Hu, and Liangji Fang.
\newblock Edgestereo: A context integrated residual pyramid network for stereo
  matching.
\newblock {\em arXiv}, 2018.

\bibitem{szeliski2008comparative}
Richard Szeliski, Ramin Zabih, Daniel Scharstein, Olga Veksler, Vladimir
  Kolmogorov, Aseem Agarwala, Marshall Tappen, and Carsten Rother.
\newblock A comparative study of energy minimization methods for markov random
  fields with smoothness-based priors.
\newblock {\em TPAMI}, 2008.

\bibitem{daisy}
Engin Tola, Vincent Lepetit, and Pascal Fua.
\newblock Daisy: An efficient dense descriptor applied to wide-baseline stereo.
\newblock {\em IEEE TPAMI}, 32(5):815--830, 2010.

\bibitem{Tonioni_2019_CVPR}
Alessio Tonioni, Fabio Tosi, Matteo Poggi, Stefano Mattoccia, and Luigi
  Di~Stefano.
\newblock Real-time self-adaptive deep stereo.
\newblock In {\em CVPR}, 2019.

\bibitem{NIPS2018_7828}
Stepan Tulyakov, Anton Ivanov, and Fran\c{c}ois Fleuret.
\newblock Practical deep stereo (pds): Toward applications-friendly deep stereo
  matching.
\newblock In {\em NEURIPS}. 2018.

\bibitem{yamaguchi2014efficient}
Koichiro Yamaguchi, David McAllester, and Raquel Urtasun.
\newblock Efficient joint segmentation, occlusion labeling, stereo and flow
  estimation.
\newblock In {\em ECCV}, 2014.

\bibitem{yang2018segstereo}
Guorun Yang, Hengshuang Zhao, Jianping Shi, Zhidong Deng, and Jiaya Jia.
\newblock Segstereo: Exploiting semantic information for disparity estimation.
\newblock In {\em ECCV}, 2018.

\bibitem{sad}
Qingxiong Yang, Liang Wang, Ruigang Yang, Henrik Stew{\'e}nius, and David
  Nist{\'e}r.
\newblock Stereo matching with color-weighted correlation, hierarchical belief
  propagation, and occlusion handling.
\newblock {\em IEEE TPAMI}, 2009.

\bibitem{yin2018hierarchical}
Zhichao Yin, Trevor Darrell, and Fisher Yu.
\newblock Hierarchical discrete distribution decomposition for match density
  estimation.
\newblock 2019.

\bibitem{ncc}
Jae-Chern Yoo and Tae~Hee Han.
\newblock Fast normalized cross-correlation.
\newblock {\em CSSP}, 2009.

\bibitem{zabih1994non}
Ramin Zabih and John Woodfill.
\newblock Non-parametric local transforms for computing visual correspondence.
\newblock In {\em ECCV}, 1994.

\bibitem{zagoruyko2015learning}
Sergey Zagoruyko and Nikos Komodakis.
\newblock Learning to compare image patches via convolutional neural networks.
\newblock In {\em CVPR}, 2015.

\bibitem{zbontar2016stereo}
Jure Zbontar and Yann LeCun.
\newblock Stereo matching by training a convolutional neural network to compare
  image patches.
\newblock {\em JMLR}, 2016.

\bibitem{zhao2017pyramid}
Hengshuang Zhao, Jianping Shi, Xiaojuan Qi, Xiaogang Wang, and Jiaya Jia.
\newblock Pyramid scene parsing network.
\newblock In {\em CVPR}, 2017.

\end{thebibliography}
}

\end{document}